\documentclass[10pt,journal,compsoc]{IEEEtran}

\usepackage{times}
\usepackage{epsfig}
\usepackage{graphicx}
\usepackage{amsmath}
\usepackage{amssymb}
\usepackage{overpic}
\usepackage{cleveref}
\usepackage{tabularx}
\usepackage{nicefrac}
\usepackage{booktabs}
\usepackage{bm}
\usepackage{multirow}
\usepackage{mathtools}
\usepackage{amsmath}

\usepackage{amssymb}
\usepackage{amsfonts}
\usepackage{amsopn}
\usepackage{graphicx} % Necessary to use \scalebox
\usepackage{textcomp}
\usepackage{xfrac}
\usepackage{bbm}
\usepackage{overpic}
\usepackage{subfig}
\usepackage{wrapfig}
\usepackage[ruled,vlined,linesnumbered]{algorithm2e}
\usepackage{multirow}
\usepackage{paralist}
\usepackage{makecell}

\usepackage{color}
\definecolor{green}{rgb}{0, 0.5, 0}
\definecolor{orange}{rgb}{0.8, 0.6, 0.2}
\definecolor{red}{rgb}{1.0, 0.0, 0.0}
\definecolor{teal}{rgb}{0.0, 0.4, 0.4}
\definecolor{purple}{rgb}{0.65,0,0.65}
\definecolor{saffron}{rgb}{0.95,0.75,0.2}
\definecolor{turquoise}{rgb}{0.0,0.5,0.5}
\definecolor{brown}{rgb}{0.5, 0.16, 0.16}

\usepackage{overpic}
\usepackage{currfile} % \currfiledir

\newlength\savedwidth

\newcommand{\ys}[1]{{\color{black}#1}}

\definecolor{lightgray}{rgb}{0.6, 0.6, 0.6}

\usepackage[normalem]{ulem}

\newcommand{\hidecomment}[1]{}
\newcommand{\pips}{\texttt{PIPS}\xspace}

\newcommand{\pipsc}{\texttt{PIPS-C}\xspace}
\newcommand{\pipss}{\texttt{PIPS-S}\xspace}

\usepackage{xspace}
\ifCLASSOPTIONcompsoc
  \usepackage[nocompress]{cite}
\else
  \usepackage{cite}
\fi

\ifCLASSINFOpdf
\else
\fi

\begin{document}
\title{Learning Positive-Incentive Point Sampling in Neural Implicit Fields for Object Pose Estimation}

\author{Yifei Shi,~\IEEEmembership{Member, IEEE},
        Boyan Wan,
        Xin Xu,~\IEEEmembership{Senior Member, IEEE},
        Kai Xu,~\IEEEmembership{Senior Member, IEEE}
\IEEEcompsocitemizethanks{\IEEEcompsocthanksitem Yifei Shi and Xin Xu are with the College of Intelligence Science and Technology, National University of Defense Technology, China.
\IEEEcompsocthanksitem Boyan Wan and Kai Xu are with the College of Computer Science, National University of Defense Technology, China.
\IEEEcompsocthanksitem Yifei Shi and Boyan Wan are joint first authors. Corresponding author: Kai Xu (kevin.kai.xu@gmail.com).
}}

\markboth{IEEE TRANSACTIONS ON PATTERN ANALYSIS AND MACHINE INTELLIGENCE}%
{Shell \MakeLowercase{\textit{et al.}}: Bare Demo of IEEEtran.cls for Computer Society Journals}

\IEEEtitleabstractindextext{%
\begin{abstract}
Learning neural implicit fields of 3D shapes is a rapidly emerging field that enables shape representation at arbitrary resolutions.
Due to the flexibility, neural implicit fields have succeeded in many research areas, including shape reconstruction, novel view image synthesis, and more recently, object pose estimation.
\ys{Neural implicit fields enable learning dense correspondences between the camera space and the object’s canonical space – including unobserved regions in camera space – significantly boosting object pose estimation performance in challenging scenarios like highly occluded objects and novel shapes.}
\ys{Despite progress, predicting canonical coordinates for unobserved camera-space regions remains challenging due to the lack of direct observational signals. This necessitates heavy reliance on the model's generalization ability, resulting in high uncertainty. Consequently, densely sampling points across the entire camera space may yield inaccurate estimations that hinder the learning process and compromise performance.}
\ys{To alleviate this problem, we propose a method combining an SO(3)-equivariant convolutional implicit network and a positive-incentive point sampling (\pips) strategy.
The SO(3)-equivariant convolutional implicit network estimates point-level attributes with SO(3)-equivariance at arbitrary query locations, demonstrating superior performance compared to most existing baselines.
The \pips strategy dynamically determines sampling locations based on the input, thereby boosting the network's accuracy and training efficiency.
The \pips strategy is implemented with a \pips estimation network which generates sparse sample points with distinctive features capable of determining all object
pose DoFs with high certainty.}
To collect the training data of the \pips estimation network, we propose to automatically generate the pseudo ground-truth with a teacher model.
Our method outperforms the state-of-the-art on three pose estimation datasets.
It achieves $0.63$ in the $5^{\circ}2$cm metric on NOCS-REAL275, $0.62$ in the $5^{\circ}5$cm metric on ShapeNet-C, and $77.3$ in the AR metric on LineMOD-O.
Notably, it demonstrates significant improvements in challenging scenarios, such as objects captured with unseen pose, high occlusion, novel geometry, and severe noise.
%Moreover, we provide in-depth qualitative analyses to reveal the advantages of individual components in the proposed method.
%We also demonstrate the cross-task generality of the learned sampling strategy by applying the trained \pips estimation network to other tasks with neural implicit fields, such as implicit shape reconstruction.
\end{abstract}

% Note that keywords are not normally used for peerreview papers.
\begin{IEEEkeywords}
Neural Implicit Fields, Point Sampling Strategy, Object Pose Estimation.
\end{IEEEkeywords}}

% make the title area
\maketitle

\IEEEdisplaynontitleabstractindextext

\IEEEpeerreviewmaketitle

\ifCLASSOPTIONcompsoc
\IEEEraisesectionheading{\section{Introduction}\label{sec:intro}}
\else
IEEEhowto:kopka\section{Introduction}
\label{sec:intro}
\fi

\IEEEPARstart{R}{ecent} work has made significant progress in learning neural implicit fields of 3D shapes, enabling compact and expressive 3D representation for shape reconstruction~\cite{park2019deepsdf,chen2019learning,mescheder2019occupancy} and image synthesis~\cite{mildenhall2021nerf,kerbl20233d}. By training to predict the SDFs on a set of sampled points, neural implicit fields are able to generate the continuous SDFs on untrained locations with good generality, thanks to the advances of coordinate-based networks and the strategy of training on densely sampled points which cover the whole space of interest.

%!TEX root = ../sceneparse.tex

\begin{figure}[b!]
   \begin{overpic}[width=1.0\linewidth,tics=10]{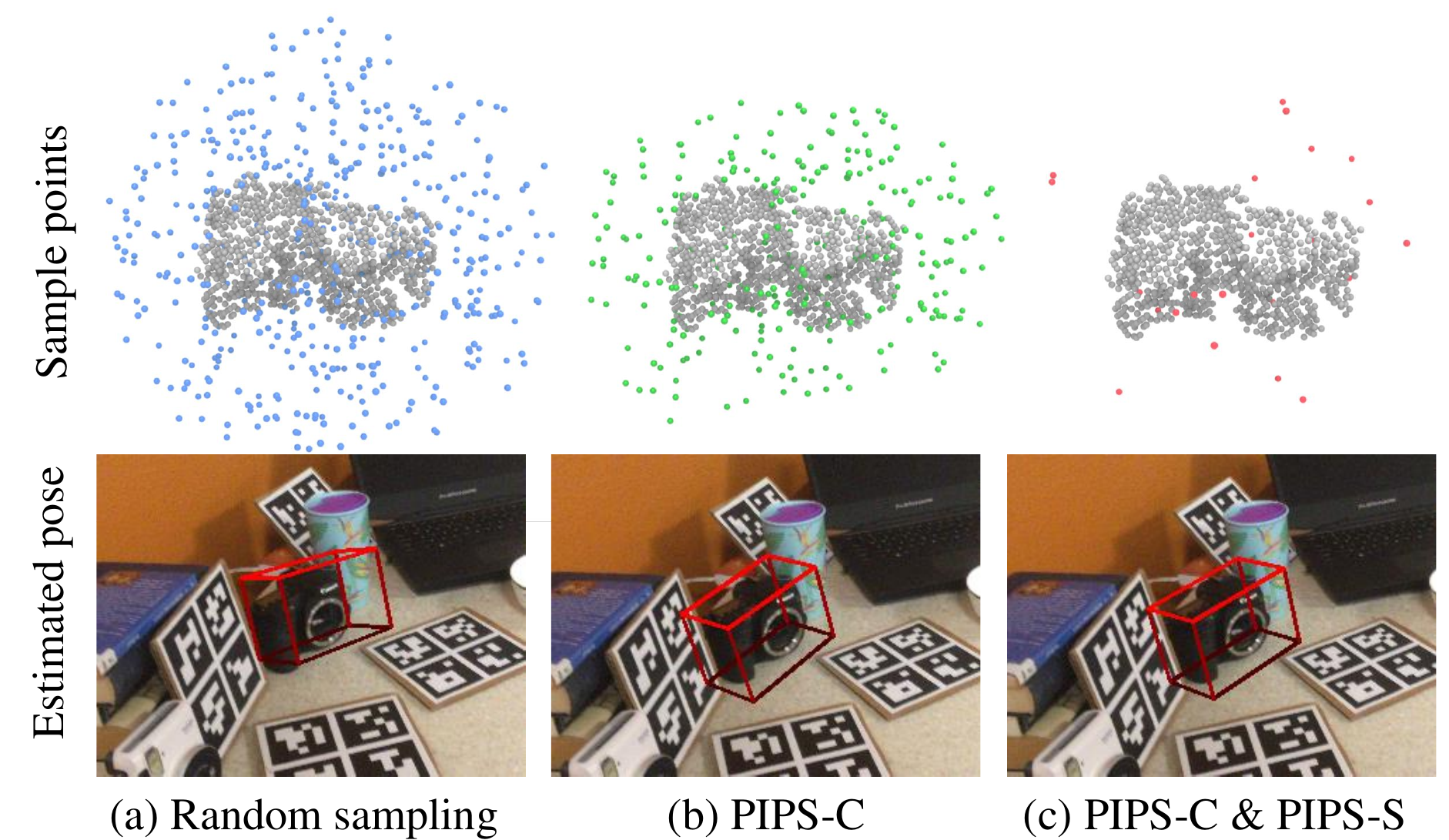}
   \end{overpic}
   \caption{We propose \pips, a data-driven approach to dynamically determine where to sample to boost the network training, achieving better performance with training on fewer sampling points, compared to (a) the random sampling baseline. \pips consists of two components: (b) positive-incentive point sampling with high estimation certainty (\pipsc) and (c) positive-incentive point sampling with high geometric stability (\pipss). }
   \label{fig:teaser}%\vspace{-12pt}
\end{figure} 

\ys{Sharing a similar idea, neural implicit fields have emerged as a powerful approach for 6D object pose estimation.
Specifically, these methods learn dense correspondences between the camera space and the object’s canonical space.
This capability extends beyond the input points to infer correspondences even for unobserved regions in the camera space~\cite{huang2022neural,agaram2022canonical,peng2022self}.
Consequently, compared to conventional methods that directly predict poses, neural implicit fields achieve significantly higher accuracy and robustness, particularly in challenging scenarios involving heavy occlusion or novel object shapes.}

%It should claim that the network learns the mapping of the camera space and the object’s coordinate space first, and then introduce the point sampling. The “unseen region” is not well defined here.

\begin{figure}[t!]
   \begin{overpic}[width=1.0\linewidth,tics=10]{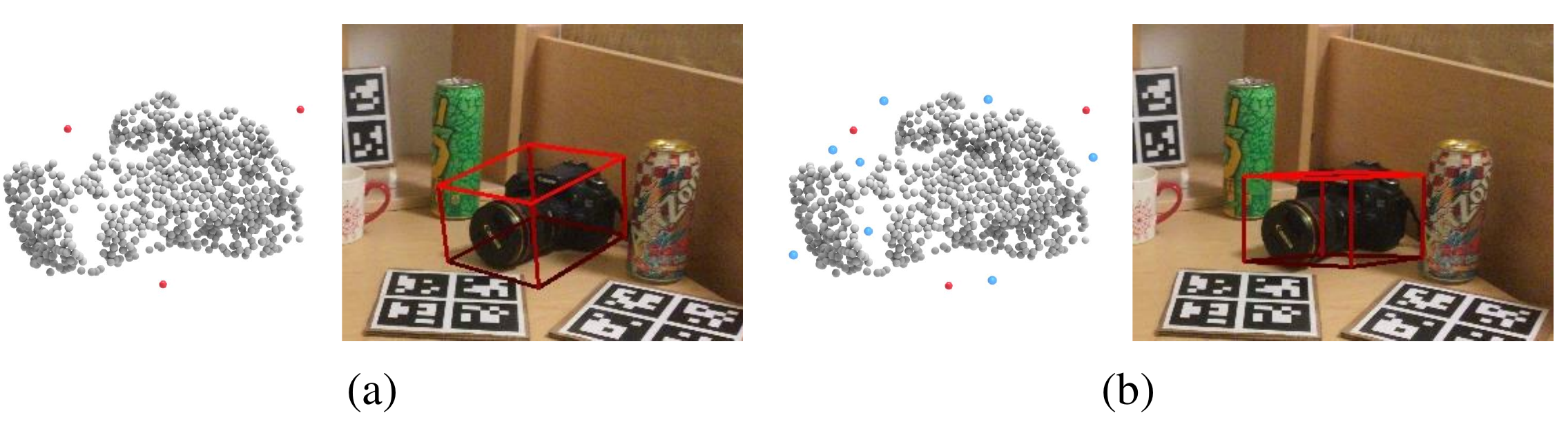}
   \end{overpic}
   \caption{(a) Point-wise canonical coordinate estimation on as few as three points (in red) is sufficient for determining all the 6-DoFs of object pose. (b) Extra voters with inaccurate point-level estimations (in blue) would degrade the performance.}
   \label{fig:stable}
\end{figure}

Despite the advantages, the whole space dense sampling is a non-optimal strategy for pose estimation.
There are two main reasons. \ys{First, the whole space dense sampling would incur hard training samples for which the network is difficult to learn, such as the sampling points from unobserved camera-space regions, due to their indistinctive features. This necessitates heavy reliance on the model’s generalization ability, resulting in high uncertainty. As a result, employing data mining to select informative, distinguishing, and learnable training samples is necessary.}
Second, unlike shape reconstruction which requires all the point-wise estimations to be as accurate as possible, pose estimation only requires accurate point-wise estimation on a limited number of locations.
As shown in Figure~\ref{fig:stable}, canonical coordinate estimation on as few as three points is sufficient for determining all the DoFs of object pose. Extra voters with inaccurate point-level estimations might degrade the overall performance.

%!TEX root = ../sceneparse.tex

To tackle this problem, in this paper, we study the problem of how to generate sample points that would boost network training.
We first provide the definition and the empirical analysis of the positive-incentive point sampling (\pips) strategy.
\ys{Specifically, we define \pips as generating sparse sample points with distinctive features capable of determining all object pose DoFs with high certainty.}
To implement this idea and achieve robust performance, we propose to learn a neural implicit field using SO(3)-equivariant convolutions, trained with sample points generated by a \pips estimation network.
%The intuitions of \pips are three-fold.
%First, the point-wise feature of the sample points should be distinctive.
%Second, all the sample points together should be able to determine all the DoFs of the object pose with high confidence.
%Third, the sampled points should be sparse in order to maintain a low computation cost.

%Training on the positive-incentive sampled points would lead to accurate pose estimation with less training cost.

%The intuition of the definition is simple: positive-incentive sampled points are excepted to maximize the information gain during network training and enable pose estimation with high confidence during network inference.
%As such, points with high and low uncertainties should be sampled during training and inference, respectively.
%To achieve this, we propose Positive-Incentive Point Sampling (PIPS), a data-driven method to dynamically determine \emph{"where to sample"} such that the training and inference of implicit pose estimation network could be best positively incentivized.

%The proposed method has two main modules: the implicit SO(3)-equivariant convolutional network and the positive-incentive point sampling (\pips).

\ys{The SO(3)-equivariant convolutional implicit network is a backbone that aggregates SO(3)-equivariant features from the input points and estimates point-level attributes at any query locations. To achieve this, the direction-independent point convolution kernels based on the vector neurons~\cite{deng2021vector} are developed, making the operation of 3D convolution SO(3)-equivariant.
By integrating the SO(3)-equivariant convolutions with a recent implicit neural network~\cite{wan2023socs}, the method outperforms most existing works of object pose estimation with implicit functions.}

\begin{figure}[t!] \centering
	\begin{overpic}[width=1.0\linewidth,tics=5]{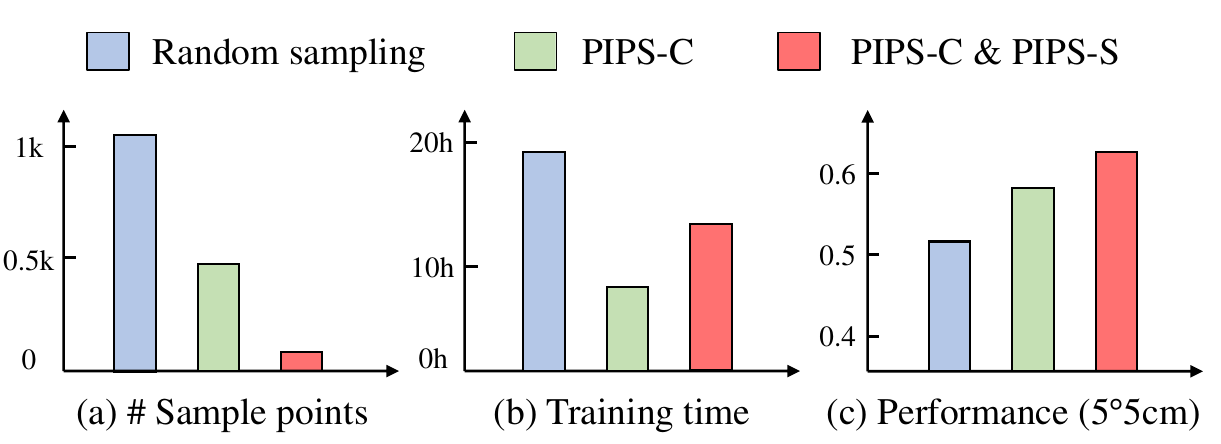}%,grid
    \end{overpic}
   \caption{The quantitative comparisons of the proposed \pipsc and \pipss to the baseline of random sampling. We see our method reduce the number of sample points and the training time while achieving better performance in object pose estimation. The experiment is conducted on the NOCS-REAL275 dataset.}
   \label{fig:statistics}
\end{figure}

To generate sample points to train the above network, we propose a \pips estimation network, a simple yet effective data-driven approach to dynamically determine where to sample to boost network training.
The \pips estimation network consists of two components: \ys{positive-incentive point sampling with high estimation certainty} (\pipsc) and positive-incentive point sampling with high geometric stability (\pipss).
The \pipsc estimation component contains a point cloud-based encoder and a volumetric grid-based decoder. It learns to generate sample points with high estimation certainty. As a result, training the SO(3)-equivariant convolutional implicit network on the \pipsc sample points would bring sufficient information gain~(Figure~\ref{fig:teaser}b).
Using all the \pipsc sample points is neither efficient nor necessary.
%According to the geometric stability analysis on point cloud registration~\cite{gelfand2003geometrically} and pose estimation~\cite{stablepose}, a minimum of three 3D points would be geometrically stable to determine the full 6DoFs of the object pose.
%Inspired by these works,
To solve this problem, the \pipss estimation component further selects the sparse and geometrically stable subsets from the above sample points~(Figure~\ref{fig:teaser}c).
The \pipss estimation component is implemented with an attentional gating module trained by the Gumbel-Softmax trick.
A stability loss function and a sparsity loss function are applied to optimize the sample points with high geometric stability while keeping the sample points sparse.
As reported in Figure~\ref{fig:statistics}, the proposed \pips greatly reduces the number of sample points and the training time while achieving better performance in pose estimation.

%!TEX root = ../sceneparse.tex

%Based on that, a \pipss estimation component is proposed to further select sparse and geometrically stable pointset, further reducing the training cost while maintaining the overall accuracy.

The \pips estimation network is trained in a knowledge distillation manner, where a sophisticated teacher model is first trained to generate the pseudo ground-truth and a student model (i.e. the \pips estimation network) is then optimized to mimic the teacher model.
%We train the teacher model with a mutual adversarial loss function that automatically generates the per-point uncertainties. The per-point uncertainties are utilized as the pseudo ground-truth in the training of the \pips estimation network.
Interestingly, we found that using a small set of pseudo ground-truth is sufficient for the \pips estimation network to generate meaningful and repeatable positive incentive points.

%We see this as analogous to knowledge distillation, i.e. transferring the knowledge from a large unwieldy model to a single smaller model that can be practically deployed with better efficiency.

%we develop \pipsc that outputs sample points with high feature certainty and consequently inducing large information gain during the training of the implicit SO(3)-equivariant convolutional network(Figure~\ref{fig:teaser} b).

% 用gpt的那些说法
% loss function包括：stability loss, canonical coordinate estimation loss

%(describe the geometric stable analysis, how to design the network, what are the multiple loss functions).%突出 theory
%(An approach with probability distribution output is proposed to handle the multiple possible outputs. LLM.) %突出问题描述和LLM

Experiments demonstrate that the proposed method outperforms the state-of-the-art on three datasets of pose estimation.
It achieves $0.63$ in the $5^{\circ}2$cm metric on NOCS-REAL275, $0.62$ in the $5^{\circ}5$cm metric on ShapeNet-C, and $77.3$ in the AR metric on LineMOD-O.
%An example of the generated sample points is illustrated in Figure~\ref{fig:teaser}.
It has better performances in various challenging scenarios, such as objects captured with unseen pose, high occlusion, novel geometry, and severe noise.
Moreover, we provide in-depth qualitative analyses of the learned sampling strategy.
We also demonstrate the cross-task generality of the learned sampling strategy by applying the trained \pips estimation network to other tasks with neural implicit fields, such as implicit shape reconstruction.

%First, an implicit fields-based pose estimation network with a dual attentional point sampling unit is present to dynamically generate the positive-incentive sampled points during training and inference, separately. The network is trained in an adversarial-synergy manner that maximizes the information gain during training and generates pose estimation results with high confidence during inference.

%Second, to generate instance-specific sampling points under moderate occlusion, the attentional point sampling unit is coupled with a pre-trained completion network, facilitating the feature aggregation in the unseen regions. Besides, several regularization modules are utilized to encourage sparse and dispersive distribution of the sampling points.
% 不仅取决于the geometry of input points, 还取决于 the complete shape，所以要把 completion as axillary task。可以实比较粗的感知，不需要特别精细
% regularization -> distribution/distance between points, number of points

%Third, a feature propagation method.
% 1) rotation-invariant network for feature aggregation. transfer absolute coordinate to relative coordinate. 2) selecting the k-means centers in the feature space.

%Fourth, training with the Gumbel-Softmax trick and RANSAC/voting differentiable algorithm, the network is trained in an end-to-end manner.

%Fifth, meta-learning.

% 4） 我们的solution：tranin和test到底需要怎么样的采样（30m）
% 5) 实验：1.说明问题的必要性；2.
% 必要性实验: 稠密训练，1）随机遮住一些点测试，发现性能没有明显下降；2）有针对性地遮掉一些点测试，发现性能反而增强.

In summary, we make the following contributions:
\begin{itemize}
\item We propose the idea of detecting positive-incentive sample points for neural implicit fields, that improve the accuracy and training efficiency.
\item \ys{We propose an SO(3)-equivariant convolutional implicit network to estimate point-level attributes, achieving better performance compared to most existing implicit neural fields in pose estimation.}
\item We develop the \pips estimation network, including a \pipsc estimation component and a \pipss estimation component, to generate sample points with high estimation certainty and high geometric stability, respectively.
\item Our method achieves state-of-the-art performance on three datasets of pose estimation. In particular, it performs well in various challenging scenarios.
\end{itemize}

\section{Related work}
\label{sec:related}
\noindent\textbf{Neural Implicit Representation of 3D Shapes.}\
Many existing works have investigated implicitly representing 3D shapes with continuous and memory-efficient implicit fields that map (x,y,z) coordinates to signed distance fields~\cite{park2019deepsdf,michalkiewicz2019implicit} or occupancy functions~\cite{chen2019learning,mescheder2019occupancy}, implemented by neural networks.
These neural implicit fields of 3D shapes allow high-quality surface representation with any resolution, facilitating the application of them in shape reconstruction.
Combined with advanced network architectures~\cite{peng2020convolutional,gropp2020implicit,takikawa2021neural} and training schemes~\cite{jiang2020local,chabra2020deep}, neural implicit fields are not limited to single objects with simple geometry but also could scale to large-scale complicated scenes.
In addition to representing geometry, recent works have explored encoding appearance with neural implicit fields~\cite{mildenhall2021nerf,barron2022mip,barron2021mip,yu2021pixelnerf}, achieving state-of-the-art visual quality of novel view image synthesis.
Despite the progress, representing 3D shapes with neural implicit fields usually requires training on densely sampled points, some of which are less informative and would increase the computational cost and lead to less accurate estimation.

\noindent\textbf{Neural Implicit Fields for Pose Estimation.}\
While most of the research in neural implicit fields focuses on shape reconstruction and view synthesis, some other methods adopt implicit fields for object pose estimation~\cite{peng2022self,agaram2022canonical}.
A conventional and straightforward solution is to reconstruct the object surface and estimate its pose simultaneously~\cite{bruns2022sdfest,pavllo2022shape,li2022generative,wen2022disp6d} so that the two tasks could boost each other.
For example, ShAPO~\cite{irshad2022shapo} jointly predicts object shape, pose, and size in a single-shot network.
DISP6D~\cite{wen2022disp6d} disentangles the latent representation of shape and pose into two sub-spaces, improving the scalability and generality.
With the mechanism of representing complex 3D geometry from a set of RGB images, Neural Radiance Fields (NeRF) is also applicable to optimize the object pose w.r.t a singe-view image.
For example, iNeRF~\cite{yen2021inerf} uses gradient descent to minimize the residual between pixels rendered from a pre-trained NeRF and pixels in an observed image.
NeRF-Pose~\cite{li2022nerf} first reconstructs the object from multiple views in the form of a neural implicit representation and then regresses the object pose by predicting pixel-wise 2D-3D correspondences between images and the reconstructed model.
Unlike the traditional correspondence-based methods which predict 3D object coordinates merely at observed pixels in the input image, Huang et al.~\cite{huang2022neural} predicts canonical coordinates at any sampled 3D in the camera frustum, generating continuous neural implicit fields of canonical coordinates for instance-level pose estimation.
Wan et al.~\cite{wan2023socs} extend the idea of dense per-point estimation to category-level pose estimation by proposing a semantically-aware canonical space and a transformer-based feature propagation module.
Our method is inspired by the previous works of dense per-point estimation. However, it generates positive-incentive sample points with a learning-based method which could improve the training efficiency of the implicit neural networks.

\noindent\textbf{Point Sampling Strategy in Neural Implicit Fields.}\
Training neural implicit fields for 3D shapes is challenging and time-consuming as it requires large sample counts to cover the region both inside and outside the surface, especially for 3D shapes with complex geometry.
Various sampling strategies have been adopted in neural implicit fields to achieve better training efficiency.
For 3D reconstruction, uniform sampling and near-surface sampling that selects a certain number of points in the boundary space or near the underlying surface, are widely used in a large number of previous works.
Some works adopt a combination of them to achieve balanced training~\cite{qian2023impdet}. %[ImpDet: Exploring Implicit Fields for 3D Object Detection]
Xu et al.~\cite{xu2020ladybird} propose a farthest point sampling algorithm resulting in a fast network. %[Ladybird: Quasi-Monte Carlo Sampling for Deep Implicit Field Based 3D Reconstruction with Symmetry]
Several adaptive point sampling strategies are developed to find the hard training points, allowing faster convergence and accurate representation of geometry details~\cite{yifan2021iso,jin2022adaptive,yariv2021volume}.
%[Iso-Points: Optimizing Neural Implicit Surfaces with Hybrid Representations]
%[Adaptive Points Sampling for Implicit Field Reconstruction of IndustrialDigital Twin]
% [Volume Rendering of Neural Implicit Surfaces]
There are also a bunch of works that focus on developing sophisticated point sampling on NeRFs.
Mildenhall et al.~\cite{mildenhall2021nerf} utilize a hierarchical sampling procedure by training an extra coarse network, allowing efficient network training. %[NeRF: Representing Scenes as Neural Radiance Fields for View Synthesis]
Li et al.~\cite{li2022nerfacc} and Sun et al.~\cite{sun2022neural} discretize the scene into voxels and compute the importance of each voxel to the rendered image. Therefore, the method can skip the invalid areas and avoid unnecessary computation.
%[Nerfacc: A general nerf acceleration toolbox]
%[Neural 3d reconstruction in the wild]
Training neural implicit fields to directly sample in an end-to-end manner is another direction~\cite{lindell2021autoint,piala2021terminerf,barron2022mip,kurz2022adanerf}. These methods require additional time to train the sampling network and would not generalize well in unseen scenarios.
%[Autoint: Automatic integration for fast neural volume rendering]
%[Terminerf: Ray termination prediction for efficient neural rendering]
%[Mip-nerf 360: Unbounded anti-aliased neural radiance fields]
%[Adanerf: Adaptive sampling for real-time rendering of neural radiance fields]
Our method is relevant to those methods. However, it is designed specifically for pose estimation from a novel prospect and can be generalized in other relevant tasks.
%#都是关于预测surface的，没有考虑点特征的certainty，也无法扩展到pose estimation， where the gt surface is not always available

\noindent\textbf{Equivariant Network for Point Cloud Analysis.}\
The equivariance property is crucial for point cloud analysis. Various approaches were proposed to address this problem. A straightforward way is to estimate the orientation from the input so that the equivariance can be obtained. There are a bunch of existing works in this direction, including object orientation estimation~\cite{wang2019densefusion,peng2019pvnet,hodan2018bop} and local patch orientation estimation~\cite{qi2017pointnet,qi2017pointnet++,wen2020edge}.
Recently, convolutions with steerable kernel bases have emerged~\cite{weiler20183d,weiler2019general,andrearczyk2019exploring,melnyk2022steerable}, using additional storage and specialized operations to guarantee equivariance for common network layers. For example, 3D Steerable CNNs adopt the convolution with steerable kernel bases~\cite{weiler20183d}. This convolution results in a rotation of the features in the feature space, inheriting equivariance from input points to output features.
Vector-based neural network is another direction~\cite{deng2021vector,jing2020learning,satorras2021n}. For example, vector neurons~\cite{deng2021vector} extend the 1D scalars to 3D vectors, enabling the mapping of SO(3) actions to the feature space.
Our method adopts the idea of vector neurons. However, we make extensions to make it applicable for the 3D graph convolution layers, greatly broadening its application scope.

% (Orientation Estimation, Vector-Based Networks) https://openaccess.thecvf.com/content/CVPR2022/papers/Luo_Equivariant_Point_Cloud_Analysis_via_Learning_Orientations_for_Message_Passing_CVPR_2022_paper.pdf
% (steerable kernels.)https://openaccess.thecvf.com/content/ICCV2021/papers/Deng_Vector_Neurons_A_General_Framework_for_SO3-Equivariant_Networks_ICCV_2021_paper.pdf 

\section{Method}
\label{sec:method}

\subsection{Definition}

%As discussed in the introduction, densely sampling points across the entire camera space may yield inaccurate estimations that hinder the learning process and compromise performance, due to the indistinctive features at certain points and the limited DoFs in object pose. To address this, we propose positive-incentive point sampling, which strategically selects points that actively facilitate network training. The sample points satisfy three criteria:
%1) Distinctive per-point features enabling high-certainty canonical coordinate prediction; 2) Inter-point complementarity for determining all object pose DoFs; 3) Sparsity to ensure training efficiency.

\ys{As discussed in the introduction, densely sampling points across the entire camera space often leads to inaccurate estimations that can hinder the learning process and compromise overall performance. This issue arises primarily due to the presence of points with indistinctive or less-informative features, which provide unreliable signals.

To tackle this challenge, we introduce positive-incentive point sampling (\pips), a strategic sampling approach designed to selectively identify points that actively contribute to and facilitate effective network training.
\pips refers to a process that predicts sample points that would positively incentivize the learning process and the overall performance of implicit neural networks.

Specifically, the sample points of \pips are required to satisfy three key criteria:
1) \emph{Distinctive features}: Each sample point must exhibit distinctive and discriminative features that enable high-certainty estimation of its canonical coordinate;
2) \emph{Sparsity}: The sample points should be sparse to maintain computational efficiency and avoid unnecessary redundancy;
3) \emph{Inter-point complementarity}: The set of sample points should collectively provide sufficient information to constrain all DoFs of the object’s pose.}

%!TEX root = ../sceneparse.tex

\begin{figure*}[t!] \centering
\begin{overpic}[width=1.0\linewidth,tics=10]{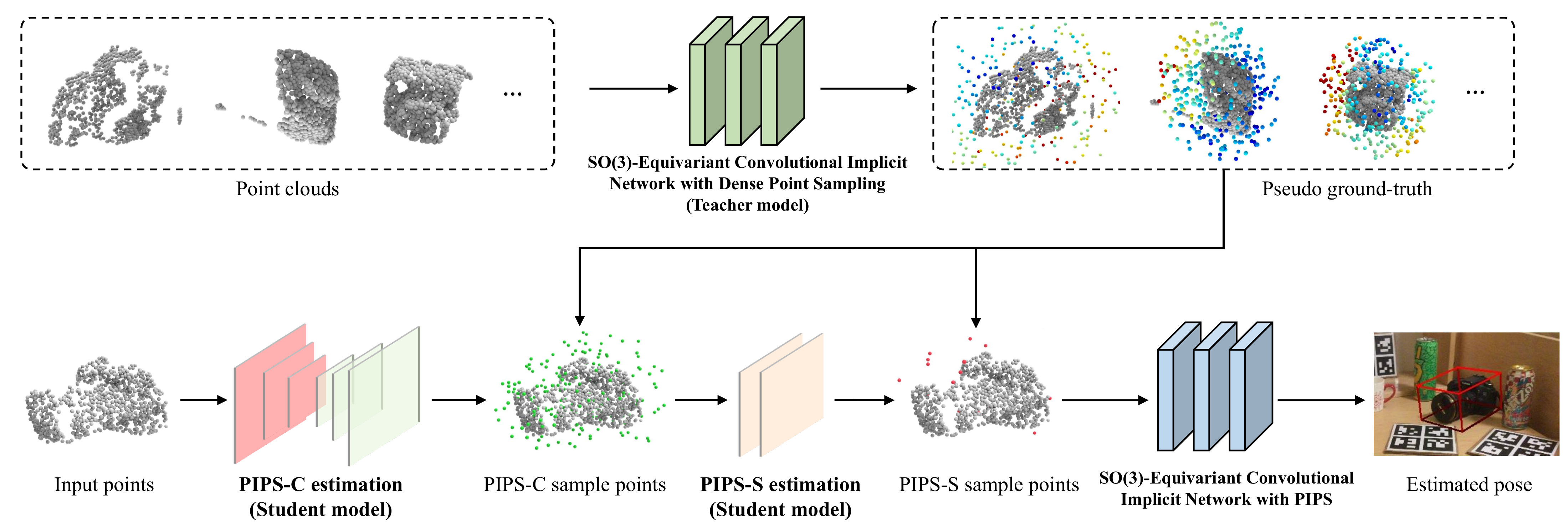}%,grid
   \end{overpic}
   \caption{\ys{Overview of the proposed method.
   First, an SO(3)-equivariant convolutional implicit network with dense point sampling (the teacher model) is optimized to generate the pseudo ground-truth. Second, the \pipsc and \pipss estimation networks (the student model) are trained based on the generated pseudo ground-truth. Third, an SO(3)-equivariant convolutional implicit network is trained with the sample points estimated by the \pips estimation network.}}
   \label{fig:training}
\end{figure*}
%\vspace{-8pt} 
%!TEX root = ../sceneparse.tex

\begin{figure}[t!]
   \begin{overpic}[width=1.0\linewidth,tics=10]{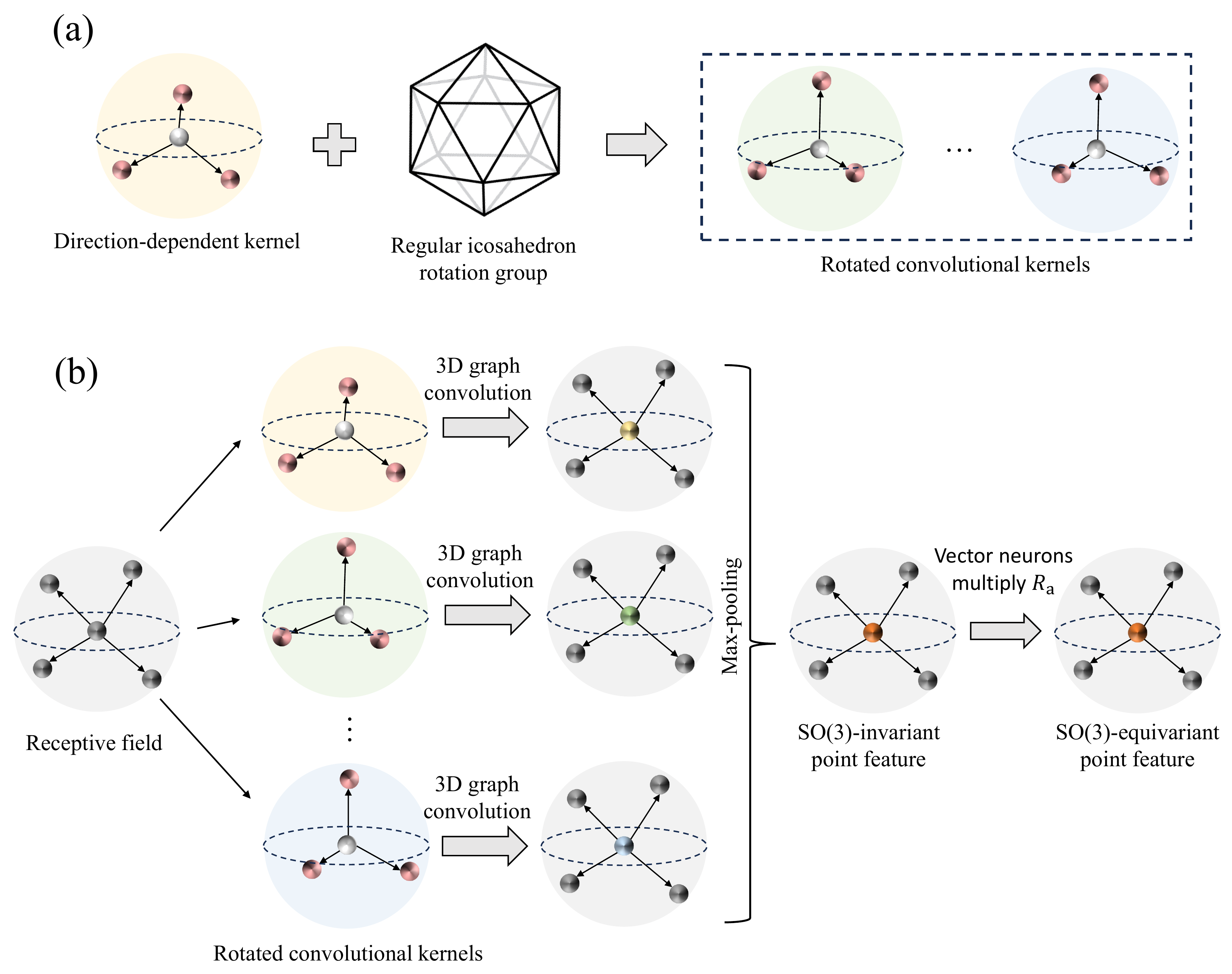}
   \end{overpic}
   \caption{(a) By rotating the 3D graph convolution kernel via a regular icosahedron rotation group, we generate a set of convolutional kernels.  (b) Point cloud convolutions with the rotated convolutional kernels make the generated features SO(3)-invariant. By generating vector neurons by the SO(3)-invariant point feature and multiplying the vector neurons with the rotation matrix $R_a$ corresponding to the $q_a\in Q$ with the highest activation, the feature becomes SO(3)-equivariant.}
   \label{fig:so3}%\vspace{-12pt}
\end{figure}

\subsection{Overview}
Our method consists of two crucial modules: the SO(3)-equivariant convolutional implicit network and the \pips estimation network.
The SO(3)-equivariant convolutional implicit network is a backbone that estimates point-wise object canonical coordinates at any query location based on the input points.
%When trained with the random point sampling strategy, it outperforms many existing baselines in pose estimation.
The \pips estimation network includes two main components: \pipsc and \pipss, which are two successive modules that boost the training of the SO(3)-equivariant convolutional implicit network.
Specifically, \pipsc generates sample points with high estimation certainty.
After that, \pipss further selects the sparse and geometrically stable subset from the sample points of \pipsc, resulting in a more parsimonious pointset.
\ys{An overview of the method is illustrated in Figure~\ref{fig:training}.}

%During training, maximizing I(pose,T') equals to select the trainable and informative T'. Not too easy and not too hard.
% during inference, maximizing I(pose,T') equals to select the crucial T'. Confident and informative in terms of determining the pose.

% mutual information entropy between I(pose,T')=H(pose)-H(pose/T')
% T' is a subset of T
% 𝐻(.) is the information entropy， 𝐻(.|.) is the conditional entropy，即知道了T'之后，确定pose的complexity

% The problem comes to finding a T' that minimize: H(pose/T')=�H(pose)−�I(pose,T'). Suppose 𝐻(�pose) is fixed, the optimization equals to maximize: I(pose,T').

% If I(pose,T')>0, 𝑇' is positive-incentive samples, else it is negative samples

% 理论上说，任意点都是Positive-incentive Point， 但是在实际任务中，从学习的角度说，有的点太难，没法训练得很准确，有的点过于重简单，没有information gain。另外，也不需要那么多点，继续增加点不会增加pose精度。
% 因此，我们定义：
% during training, maximizing I(pose,T') equals to select the trainable and informative T'. Not too easy and not too hard.
% during inference, maximizing I(pose,T') equals to select the crucial T'. Confident and informative in terms of determining the pose.

\subsection{SO(3)-Equivariant Convolutional Implicit Network}
\label{sec:backbone}
%We first describe the implicit neural network that allows point-wise estimation with feature SO(3)-equivariance.
\ys{While previous non-equivariant neural networks could produce satisfactory point-level predictions, their training typically required data augmentation to ensure the training data represented a sufficiently diverse range of poses over the SO(3) group.}
SO(3)-equivariant neural networks reduce the model complexity, accelerating the training process and leading to more robust predictions.
Recently, various methods have been developed to provide SO(3)-equivariance to basic neural network layers, but few can incorporate sophisticated layers, such as 3D convolution layers.
To facilitate an effective feature extraction for 3D convolution layers, we propose an SO(3)-equivariant 3D graph convolution layer.

\ys{Implementing 3D convolution with SO(3)-equivariance presents several challenges.
First, 3D convolutions aggregate information from local neighborhoods, as the relative positions and orientations of neighboring points change under rotation. The characteristic inherently breaks SO(3)-equivariance.
Second, implementing 3D convolution with continuous SO(3)-equivariance requires substantial memory and computation cost.
The proposed SO(3)-equivariant 3D graph convolution is a kernel-based approach that allows us to compute the convolution with a limited number of rotations, maintaining a balance between representation ability and computational cost.}

%\ys{SO(3)-Equivariant is crucial; existing SO(3)-Equivariant Network cannot incorporated with point convolution, limiting its application in real scanned data; we propose a network to achieve this, 1h}
%\ys{1) network architecture; 2) equivariance}
%\noindent\textbf{Network architecture.}\
%the existing implicit network~\cite{wan2023socs} with several feature equivariant modules, making the existing implicit network SO(3)-equivariant and improving the training efficiency.
%Specifically, we enhance the existing implicit network in the following two aspects.

%!TEX root = ../sceneparse.tex

\begin{figure*}[t!] \centering
	\begin{overpic}[width=1.0\linewidth,tics=10]{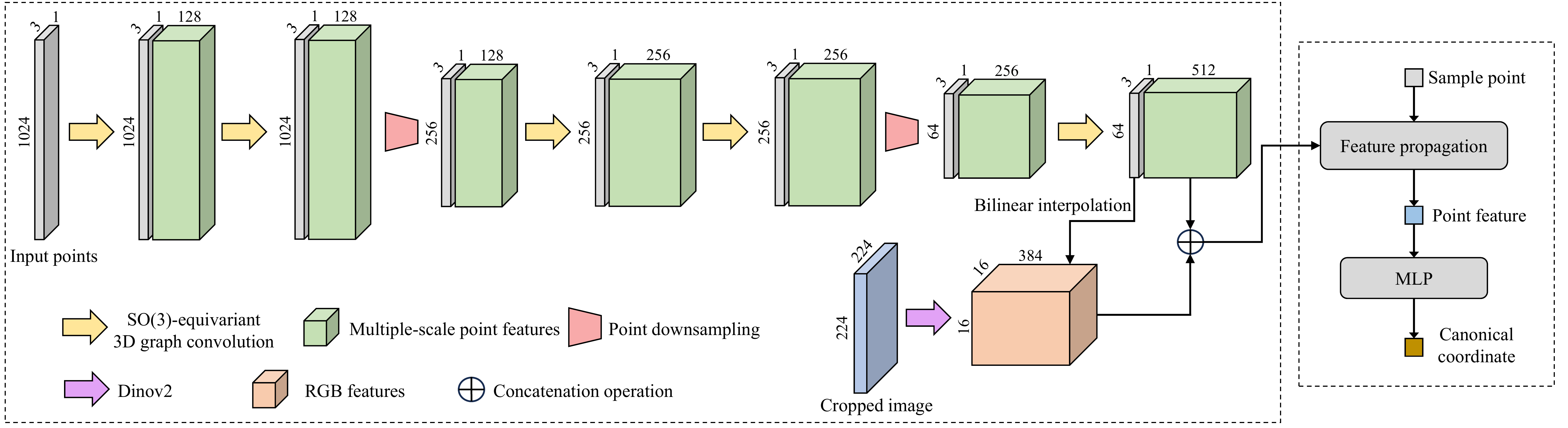}%,grid
   \end{overpic}
   \caption{Architecture of the SO(3)-equivariant convolutional implicit network. Left: Multiple SO(3)-equivariant 3D graph convolution layers and point downsampling layers are adopted to aggregate multiple-scale point features. \ys{The RGB features are extracted by DINOv2.} Right: The aggregate features are propagated to any 3D sample point via feature propagation layers, estimating the point-wise object canonical coordinate.
   }
   \label{fig:so3_pipeline}
\end{figure*} 

\subsubsection{SO(3)-equivariant 3D graph convolution layer}
The SO(3)-equivariant 3D graph convolution layer takes a point cloud $P$ as input and generates the per-point convolutional SO(3)-equivariant features.
To achieve this, we adopt the idea of vector neurons~\cite{deng2021vector} that extend each 1D scalar neuron in the network to a 3D vector.
The vanilla vector neurons allow direct mapping of rotations on several basic layer types, including the linear layer, the pooling layers, and the normalization layers.
\ys{
To enforce SO(3)-equivariance in point cloud convolution while maintaining a balance between performance and computational overhead, we extend the vector neurons to make them applicable to the 3D graph convolution layers~\cite{lin2020convolution}.
}

The process of the proposed SO(3)-equivariant 3D graph convolution is shown in Figure~\ref{fig:so3}.
Suppose the feature of each point $p_n$ in $P$ is $f(p_n) \in \mathbb{R}^{C\times3}$. The receptive field of $p_n$ is $\Pi_n^M$. $M$ is the number of points in the receptive field. The learnable 3D convolution kernel $K_U=\{k_0,k_1,k_2,...,k_U\}$ composes of one center point and $U$ support points.
$k_0=(0,0,0)$ is the center point.
$k_u$ ($u\in\{1,2,...,U\}$) are the support points in the kernel.
\ys{The 3D graph convolution is:
\begin{equation}\label{eq:conv}
\text{Conv}(\Pi_n^M,K_U)=\left \langle f(p_n),w(k_0)\right \rangle+\Theta,
\end{equation}
where $w(k_0) \in \mathbb{R}^{C\times3}$ is the weight matrix of $k_0$.
$\left \langle,\right \rangle$ is the feature distance.
$\Theta$ measures the similarity between $\Pi_n^M$ and the support points in $K_U$.
Specifically, $\left \langle,\right \rangle$ is the sum of the inner-product operation of each 1D neuron in $f(p_n)$ and $w(k_0)$.
%As a result, $\left \langle f(p_n),w(k_0)\right \rangle$ represents the similarity between the feature of $p_n$ and the kernel center.
$\Theta$ is computed as:}
\begin{equation}\label{eq:A}
\Theta=\max_{q\in Q}\sum_{u=1}^{U}\max_{m\in(1,M)}\text{sim}(p_m,k_u^q).
\end{equation}
$Q$ is a rotation group. We implement it with the regular icosahedron rotation group. $k_u^q$ is the support point rotated by $q\in Q$. $p_m$ ($m\in\{1,2,...,M\}$) is point in the receptive field $\Pi_n^M$.
$\text{sim}(\cdot)$ is the similarity between the feature of $p_m$ and the weight of $k_u^q$:
\begin{equation}\label{eq:sim}
\text{sim}(p_m,k_u^q)=\left \langle f(p_m),w(k_u^q)\right \rangle \frac{\left \langle d_{m,n},k_u^q \right \rangle}{\|d_{m,n}\|\cdot\|k_u^q\|},
\end{equation}
where $d_{m,n}$ denotes the direction from point $p_n$ to point $p_m$.
Note that $k_u$ ($u\in\{1,2,...,U\}$), $w(k_0)$, and $w(k_u)$ ($u\in\{1,2,...,U\}$) are learnable elements which are optimized during network training.

\ys{As Equation~\ref{eq:A} selects the rotation $q_\text{activation}\in Q$ with the highest activation, the operation in Equation~\ref{eq:conv} is SO(3)-invariant.
To further make the 3D graph convolution SO(3)-equivariant, we generate a 3D vector neuron by duplicating the output of Equation~\ref{eq:conv}. We then multiply the 3D vector neuron with the rotation matrix $R_\text{activation}$ corresponding to the rotation $q_\text{activation}$, which encodes the directional information into the output.}

\subsubsection{Network architecture for pose estimation}
The architecture of the SO(3)-equivariant convolutional implicit network is shown in Figure~\ref{fig:so3_pipeline}. To train a network to output point-wise object canonical coordinates from which the object pose could be estimated, we apply multiple SO(3)-equivariant 3D graph convolution layers to aggregate multiple-scale point features.
To be specific, the network contains five SO(3)-equivariant 3D graph convolution layers and two point downsampling layers.
\ys{To leverage the information in the input RGB image, we crop the image to make it only contain the target object and feed it into the DINOv2~\cite{oquab2023dinov2} to extract the RGB features.
The downsampled point cloud is projected onto the RGB feature maps to fetch its RGB features with a bilinear interpolation.
The point features, as well as the RGB features, are concatenated to predict the point-wise object canonical coordinates of the input points.}
Moreover, to facilitate the object canonical coordinate estimation at any 3D sample points, we adopt the feature propagation layers in~\cite{wan2023socs} to estimate features at query locations.
Note that, all the 1D scaler neurons in the feature propagation layers are represented as 3D vectors to guarantee SO(3)-equivariance.
\ys{The propagated features are then converted to SO(3)-invariant features with the invariant layer in~\cite{deng2021vector} and used to predict the object's canonical coordinates with an MLP.}
The network is trained on the sample points with the mean squared error loss function.
\ys{The object pose is then estimated with the method in~\cite{wang2019normalized}.}

\subsection{\pips Estimation Network}
Having introduced the SO(3)-equivariant convolutional implicit network, we then describe the \pips estimation network that generates sampling points to positively incentivize the training of it.
The \pips estimation network is trained in a knowledge distillation manner, where a sophisticated teacher model is first trained to generate the pseudo ground-truths and the \pips estimation network is then optimized to learn from those.
In the following, we first describe the process of generating the pseudo ground-truth and then elaborate on the network architecture of the \pips estimation network.

%To achieve this, we propose the \pips estimation network, a lightweight sample point generation network containing a point cloud-based encoder and a volumetric grid-based decoder.
%The \pips estimation network is trained in a knowledge distillation manner, where a sophisticated teacher model first generates the pseudo ground-truth and a student model, i.e. the \pips estimation network, is then optimized to mimic the teacher model.

%!TEX root = ../sceneparse.tex

\begin{figure*}[t!] \centering
	\begin{overpic}[width=1.0\linewidth,tics=10]{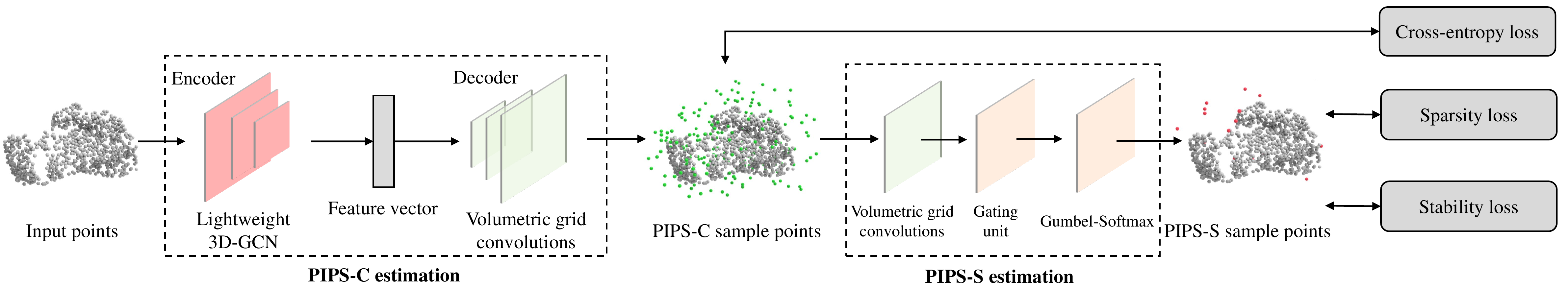}%,grid
   \end{overpic}
   \caption{The network architecture of the \pips estimation network. The network takes a point cloud as input, embeds it into a high-dimensional feature space with an encoder, and generates the \pipsc sample points with a decoder. A gating unit and a Gumbel-Softmax module are applied to generate the \pipss sample points. Multiple loss functions are adopted to train the network.
   }
   \label{fig:pips_pipeline}
\end{figure*} 

%\noindent\textbf{Generating pseudo ground-truth.}\
\subsubsection{Generating pseudo ground-truth}
\label{sec:generate_gt}
%We first elaborate on how to optimize the teacher model to generate the pseudo ground-truth.
Manually annotating the positive-incentive sampled points is infeasible due to the infinite potential sample locations and the lack of explicit labeling rules.
Hence, we propose to achieve this by optimizing a neural network that automatically generates those points.
The neural network (i.e. the teacher model) is implemented with the SO(3)-equivariant convolutional implicit network with an extra point-wise uncertainty estimation mechanism.
Specifically, we adopt the dense sampling strategy which trains the teacher model on random locations $\Phi$ spread near the input point cloud.
For each $\phi \in \mathbb{R}^{3}$ in $\Phi$, we not only estimate its object canonical coordinate but also add an output head to estimate a Gaussian distribution $(x_\phi,\sigma_\phi)$ to represent the uncertainty, where $x_\phi\in \mathbb{R}^{3}$ and $\sigma_\phi\in \mathbb{R}$.
The network could be trained with a point-level adversarial loss function:
\begin{equation}\label{eq:adv}
\mathcal{L}_\text{ADV}=\frac{1}{ \sigma_\phi^2}\Vert x_\phi-\hat{x}_\phi\Vert^2+\log \sigma_\phi^2,
\end{equation}
where $\hat{x}_\phi$ is the ground-truth object canonical coordinate.
\ys{The intuition of the loss function is to encourage accurate object canonical coordinate estimation when the point feature is certain~\cite{kendall2017uncertainties}.}
In such cases, the point is positive-incentive since it would bring sufficient information gain by training on it.
Otherwise, it encourages high variance $\sigma_\phi$, implying that it is not positive-incentive.

Despite the ability of the above loss function to filter the less informative sample points, we found the point-level uncertainty estimation should be more fine-grained.
%For example, the estimations near a planar surface should be certain along the direction of the plane normal and uncertain along the other directions.
As such, the variance $\sigma_\phi$ should be anisotropic, i.e. contains uncertainties along each direction.

In our method, instead of estimating a scalar variance $\sigma$, the network predicts a scaling matrix $S\in \mathbb{R}^{3\times 3}$ and a quaternion $q$ that represents a rotation $R\in \mathbb{R}^{3\times 3}$.
The rotated covariance matrix is represented as $\Sigma_\phi=RSS^\text{T}R^\text{T}$.
Considering the anisotropic variance, the point-level adversarial loss function could be computed by the Kullback-Leibler divergence for multivariate Gaussian distributions~\cite{duchi2007derivations}:
\begin{equation}\label{eq:adv2}
\begin{aligned}
\mathcal{L}_\text{ADV}=&\left(x_\phi-\hat{x}_\phi\right)^\text{T} \Sigma_\phi^{-1}\left(x_\phi-\hat{x}_\phi\right)\\
&+\ln {\left|\Sigma_\phi\right|}+\operatorname{tr}\left(\Sigma_\phi^{-1} \right),
\end{aligned}
\end{equation}
where $\operatorname{tr}(\cdot)$ is the trace of the matrix.

% network: generate per-point orientated gaussian. (possibly with some detailed description)
%根据论文Geometrically Stable Sampling for the ICP Algorithm中的结论，上式可以写成：
%$$\left[\Delta \mathbf{r}^T \Delta \mathbf{t}^T\right] C\left[\begin{array}{l}\Delta \mathbf{r} \\ \Delta \mathbf{t}\end{array}\right]$$
%其中$\Delta \mathbf{r}$是旋转矩阵的旋转向量形式，令$p_i= S_{i}^{-1}R_{\sigma {i}}$
%则
%$$C=\left[\begin{array}{ccc}\mathbf{y}_1 \times p_1 & \ldots & \mathbf{y}_k \times p_k\\ \mathbf{p}_1 & \cdots & \mathbf{p}_k\end{array}\right]\left[\begin{array}{cc}\left(\mathbf{y}_1 \times \mathbf{p}_1\right)^T & \mathbf{p}_1^T \\ \cdots & \ldots \\ \left(\mathbf{y}_k \times \mathbf{p}_k\right)^T & \mathbf{p}_k^T\end{array}\right]$$

%如果C有一个特征值特别小，说明存在一个方向$[\Delta r, \Delta t]$, 当所有采样点沿着这个方向移动时，误差变化较小，所以这些采样点不稳定

We train the teacher model until convergence.
For each object, we label the points with $\operatorname{tr}(SS^\text{T}) \textless \omega$ as positive, and vice versa. $\omega$ is the threshold.
To facilitate the training of \pips estimation network, the generated labels by the teacher model at random sample points $\Phi$ are converted into the labels of the volumetric grids. The label of any center point in the voxels is computed by voting considering all the sample points in this voxel.

We dub the generated labels \emph{pseudo ground-truth}, as they are generated by an auxiliary task, cannot be rigorously defined, and might be inaccurate.
Nevertheless, we found the generated samples are not only geometrically meaningful but also can boost the performance.
Please refer to the experiment section for the quantitative and qualitative comparisons.

%We found training with a dense sampling strategy will bring point-wise completion on network computation resources.
%\noindent\textbf{\pipsc estimation network.}\
\subsubsection{Network architecture}
As shown in Figure~\ref{fig:pips_pipeline}, the \pips estimation network takes the point cloud $P$ as well as the corresponding RGB image as input and outputs positive-incentive sample points.
We divide the network into two sequential components \pipsc and \pipss that generate \ys{sample points with high estimation certainty} and sample points with high geometric stability, respectively.

\noindent\textbf{\pipsc estimation.}\
The \pipsc estimation component takes $P$ as input, embeds it into a high-dimensional feature space $\mathbb{R}^{d}$ with a point cloud based encoder, and generates sample points with a volumetric grid based decoder.
Each voxel in the output volumetric grid contains a label indicating whether it is a valid positive-incentive sample point.
%To make the \pipsc estimation network lightweight and effective,
We adopt 3D-GCN~\cite{lin2020convolution} as the encoder and the convolutional occupancy networks as the decoder~\cite{peng2020convolutional}.
\ys{In the 3D-GCN, to leverage the color information in the input RGB image, we crop the image to make it only contain the target object and feed it into the DINOv2~\cite{oquab2023dinov2} to extract the RGB features.
The downsampled point cloud is projected onto the RGB feature maps to fetch its RGB features.}
The output volumetric grid $V$ shares the same center as $P$ and includes $h^3$ voxels. We set the side length of $V$ as double of the diagonal length of the target object, for instance-level pose estimation, and double of the diagonal length of the categorical mean shape~\cite{tian2020shape}, for category-level pose estimation, respectively.

The following loss function is applied to optimize the network:
\begin{equation}\label{eq:loss1}
\mathcal{L}_\text{\pipsc}=\sum_{v\in V}\mathcal{L}_\text{CE}(o_v,\hat{o}_v),
\end{equation}
where $\mathcal{L}_\text{CE}(\cdot)$ is the cross-entropy loss. $o_v\in\{0,1\}$ is the estimated label of voxel $v$. $o_v=1$ indicates the center of the voxel is a positive-incentive sample point, and vice versa. $\hat{o}_v$ is the pseudo ground-truth, which is pre-generated in Section~\ref{sec:generate_gt}.
We denote the sample points generated by \pipsc estimation component as $P_{\pipsc}$.

%\subsubsection{Discussion}
%1. lightweight

%2. solve two importance problems: 1)The ability of feature aggregation to "unseen region" is unclear.
%Automatically selecting these region.
%2) Select the 3D location with most significant and distinctive "geometry" feature to sample.

%3. why not using attention: attention select useful features to estimate global attributes, different to our task (我们是选能够学到的，即不是太难的；attention是学习与最终任务最相关的)（在我们的问题中，每个点都和最终任务相关，所以attention没有用）

% soft loss function to deal with the "pseudo ground truth", 只用方差最大和最小的那些样本
% analysis of network inference: time cost, visualization
% 1h

% 设计图：2h

%%%%%%%%%%%%%%%%%%%%%%%%%%%%%%%%%%%%%%%%%%%%%
% how to generate these points, learn form data, share repeatable patterns. 1) a offline method to generate these points, 2) use a network to learn to predict these points, such that we can get these points online by network forward pass

%1） 在SOCS基础上做一些改进，引入旋转不变性
%2） 对于每个点，预测三个方向（x,y,z）的位置、方差和R，R代表方向
%3）选择方差小的点作为PIPS-T点，形成GT
%4）从PIPS-T GT中计算得到stable points，作为PIPS-I的GT
%5）训练PIPS-T点预测网络
%6）训练PIPS-I点预测网络

%!TEX root = ../sceneparse.tex

\begin{figure}[t!]
   \begin{overpic}[width=1.0\linewidth,tics=10]{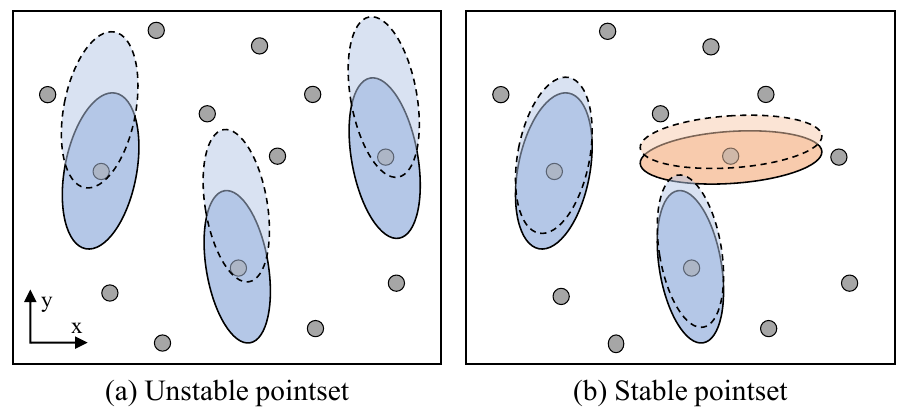}
   \end{overpic}
   \caption{\ys{A 2D example of \pipss selection. The dots are the input to the \pipss estimation component. The dots with an ellipse are the selected \pipss sample points, where the ellipse represents its anisotropic variances. (a) The pointset is unstable w.r.t. the alignment because it has a high variance along the y-axis direction, meaning that the locations along the y-axis are not well-constrained. (b) The pointset is stable w.r.t. the alignment since no DoF shows high variance for the points within the set, indicating the locations are well-constrained.}
   }
   \label{fig:pipss}
\end{figure}  

\noindent\textbf{\pipss estimation.}\
%\subsubsection{\pipss estimation}
The \pipsc estimation component generates sample points with \ys{high estimation certainty} and would boost the training of the SO(3)-equivariant convolutional implicit network.
Nevertheless, it is unnecessary to use all of them.
To further generate parsimonious sample points, we propose the \pipss estimation component.

We first mask the feature of the last layer in the \pipsc estimation component by assigning $0$ to the feature of voxels with $o_v=0$.
Taking the masked feature as input, the \pipss estimation component adopts two volumetric grid convolution layers, followed by a ReLU layer, to aggregate the feature $F_{\pipsc}$ of $P_{\pipsc}$ .

A softmax pooling layer is adopted to generate the global feature which is then concatenated with the feature vector in each voxel.
A gating unit $G$ is then applied, estimating a soft gating decisions $G^\text{soft}$ which is a mask to indicate their activations:
\begin{equation}\label{eq:gating1}\small
G^\text{soft}=\tanh[G(F_{\pipsc})+\ys{\eta}],
\end{equation}
\ys{where $\eta$ is the Gumbel noise}, we implement the gating unit $G$ with 1D MLPs.

Then, a Gumbel-Softmax module $\text{GSM}$~\cite{jang2016categorical} turns soft decisions $G^\text{soft}$ into hard decisions $G^\text{hard}\in \{0,1\}^{\Theta}$ by replacing the softmax with an argmax during the forward pass and retaining the softmax during the backward pass~\cite{kong2019pixel,verelst2020dynamic}:
\begin{equation}\label{eq:gating2}\small
G^\text{hard}=\text{GSM}(G^\text{soft}).
\end{equation}
The hard decision $G^\text{hard}$ is a binary mask that indicates which points in $P_{\pipsc}$ are activated.
The Gumbel-Softmax module provides a mechanism that outputs a binary mask in the forward pass and allows the gradient to be back-propagated.
Hence, the gating attentional unit is end-to-end trainable.
We denote the activated points selected by the gating unit $G^\text{hard}$ as $P_{\pipss}$.
The \pipss estimation component is trained with the sparsity loss function $\mathcal{L}_\text{Sparsity}$ and the stability loss function $\mathcal{L}_\text{Stability}$:
\begin{equation}\label{eq:loss2}
\mathcal{L}_\text{\pipss}=
\mathcal{L}_\text{Sparsity}+\mathcal{L}_\text{Stability}.
\end{equation}

The sparsity loss function considers the number of the activated points.
Suppose $\#P_{\pipss}$ and $\#P_{\pipsc}$ are the point number of $P_{\pipss}$ and $P_{\pipsc}$, respectively. $\mathcal{L}_\text{Sparsity}$ is computed as the KL divergence between the pre-defined empirical sparsity $g=\frac{\#P_{\pipss}}{\#P_{\pipsc}}$ and a target sparsity $\rho\in[0,1]$:
\begin{equation}\label{eq:sparsity}\small
\mathcal{L}_\text{Sparsity}=KL(\rho\Vert g)=\rho \text{log}(\frac{\rho}{g})+(1-\rho)\text{log}(\frac{1-\rho}{1-g}).
\end{equation}

\ys{The stability loss function optimizes the distribution of $P_{\pipss}$ from the perspective of geometric stability during point cloud alignment~\cite{gelfand2003geometrically,rusinkiewicz2001efficient}. In the context of aligning two point clouds (here, the input point cloud and the object point cloud in canonical space), the input point cloud is considered geometrically stable w.r.t. the alignment if all DoFs are well-constrained, meaning that no DoF exhibits high variance. Conversely, the input point cloud is deemed geometrically unstable w.r.t. the alignment if any DoF shows high variance.

A 2D example is provided in Figure~\ref{fig:pipss}. The dots represent the input to the PIPS-S estimation module. Dots marked with ellipses indicate the selected PIPS-S sample points, where each ellipse illustrates the anisotropic variance of the corresponding point. In Figure~\ref{fig:pipss}(a), the point set is geometrically unstable during alignment due to high variance along the y-axis. This implies that the vertical positions are not well-constrained, allowing the point cloud to slide along that direction. In contrast, the point set in Figure~\ref{fig:pipss}(b) is geometrically stable, as no DoF displays significant variance across the samples, indicating that all point locations are sufficiently constrained.}

%This metric also has an advantage that it allows the two surfaces to “slide” against each other in the flat and spherical regions, which do not contain enough information to fully constrain the transform. However, if too many point-pairs come from such featureless regions, the algorithm can fail to converge because of lack of constraints. In this case, the cause of poor convergence and poor final pose is the shallow error landscape that results from too much sliding. We will call geometry that does not have enough constraints for good convergence “unstable.”

%\ys{Intuitively, the key insight driving this loss is: a pointset is considered unstable during alignment if any of the DoFs exhibits high uncertainty across the entire set of points. Conversely, a stable pointset during alignment is one where all DoFs are well-constrained, meaning no DoF shows high variance for the points within the set.}

We then describe how to compute the stability in practice.
Suppose $p\in P_{\pipss}$. $x_p$ is the coordinate of $p$. $\Sigma_p=R_{p}S_{p}S_{p}^{T}R_{p}^{T}$ is the rotated covariance matrix of $p$ pre-computed in Section~\ref{sec:generate_gt}.
We compute the stability of pointset $P_{\pipss}$ under point-wise anisotropic uncertainties $\{\Sigma_p\}$ as follows.

\ys{When a perturbation transformation $[\Delta R,\Delta t]$ is applied to $P_{\pipss}$, the movement of the pointset $\Delta E_\text{\pipss}$ considering the point-wise anisotropic uncertainties is defined as:
\begin{equation}\label{eq:alignerror1}
\Delta E_\text{\pipss}=\sum_{p\in P_{\pipss}}(\Delta E_p^T \Delta E_p),
\end{equation}
where $\Delta E_p$ is the per-point movement:
\begin{equation}\label{eq:alignerror2}
\Delta E_p=S^{-1}_pR^T_p(\Delta R x_p+\Delta t-x_p),
\end{equation}
where $S_p^{-1} R_p^{T}$ incorporates directional weighting from the anisotropic uncertainty:
$R_p^{T}$ rotates the displacement into the principal-axis frame of the covariance,
and $S_p^{-1}$ scales each principal direction by the reciprocal of its standard deviation,
assigning a lower weight to directions with higher uncertainty.

\medskip
Next, we approximate Equation~(\ref{eq:alignerror1}) into the quadratic form:
\begin{equation}\label{eq:alignerror5}
	\Delta E_\text{\pipss} = [\Delta r \ \Delta t]^{T} C \, [\Delta r \ \Delta t],
\end{equation}
where $\Delta r \in \mathbb{R}^{3}$ denotes the rotation vector associated with the rotation matrix $\Delta R \in \mathbb{R}^{3\times 3}$. $C$ is the covariance matrix accumulated from $x_p$ and $\Sigma_p$.
It encodes the increase in the alignment error when the transformation is moved away from its optimum~\cite{gelfand2003geometrically}.

The relation between the rotation vector $\Delta r$ and the rotation matrix $\Delta R$ can be expressed by expanding $\Delta R$ via the Taylor series:
\begin{equation}\label{eq:alignerror3}
	\Delta R = I + \Delta r^{\wedge} + \frac{ (\Delta r^{\wedge})^{2} }{2!} + \frac{ (\Delta r^{\wedge})^{3} }{3!} + \cdots,
\end{equation}
where $\Delta r^{\wedge}$ is the skew-symmetric matrix representation of $\Delta r$.
Retaining only the first-order term in (\ref{eq:alignerror3}) and substituting into (\ref{eq:alignerror2}) gives:
\begin{equation}
	\begin{aligned}
		\Delta E_p &\approx S_p^{-1} R_p^{T} \big[ (I + \Delta r^{\wedge}) x_p - x_p + \Delta t \big] \\
		&= S_p^{-1} R_p^{T} \big[ \Delta r^{\wedge} x_p + \Delta t \big] \\
		&= S_p^{-1} R_p^{T} \big[ -x_p^{\wedge} \Delta r + \Delta t \big] \\
		&= \big[ -S_p^{-1} R_p^{T} x_p^{\wedge} \quad S_p^{-1} R_p^{T} \big]
		\begin{bmatrix} \Delta r \\ \Delta t \end{bmatrix},
	\end{aligned}
\end{equation}
where we used the property $(\Delta r)^{\wedge} x_p = -x_p^{\wedge} \Delta r$.

\medskip
Substituting the above Equation~(\ref{eq:alignerror2}) of $\Delta E_p$ into Equation~(\ref{eq:alignerror1}) yields the quadratic form in Equation~(\ref{eq:alignerror5}),
where the matrix $C \in \mathbb{R}^{6\times 6}$ is given by:
\begin{gather}
	C = C_1 C_2, \\
	C_1 =
	\begin{bmatrix}
		x_1^{\wedge} R_{1} S_1^{-1} & \cdots & x_k^{\wedge} R_{k} S_k^{-1} \\
		R_{1} S_1^{-1} & \cdots & R_{k} S_k^{-1}
	\end{bmatrix}, \\
	C_2 =
	\begin{bmatrix}
		- S_1^{-1} R_{1}^{T} x_1^{\wedge} & S_1^{-1} R_{1}^{T} \\
		\vdots & \vdots \\
		- S_k^{-1} R_{k}^{T} x_k^{\wedge} & S_k^{-1} R_{k}^{T}
	\end{bmatrix},
\end{gather}
where the sign change in $C_1$ follows from the skew-symmetric property
$\left(x_p^{\wedge}\right)^{T} = -\,x_p^{\wedge}$.
Here, $x_1, \dots, x_k$ are the coordinates of the points in $P_{\pipss}$,
and $x^{\wedge}$ denotes the skew-symmetric matrix associated with $x$.
}

The larger the pointset movement $\Delta E_\text{\pipss}$ increase, the greater the stability, since the error landscape will have a deep, well-defined minimum. On the other hand, if there are perturbation transformations that only lead to a small increase in pointset movement, the pointset will be unstable concerning one or more DoFs of the perturbation transformation.

The eigenvectors of $C$ could reflect the pointset stability. If all eigenvalues of $C$ are large, any transformation away from the minimum will result in a relatively large increase in pointset movement.
Let $\lambda_1$, ..., $\lambda_6$ be the
eigenvalues of matrix $C$. To make the pointset stable, all the eigenvalues should be large. As a result, the stability loss function penalizes small eigenvalues:

\begin{equation}\label{eq:stability}\small
\mathcal{L}_\text{Stability}=\sum_{i=1}^{6}e^{-\lambda_i}.
\end{equation}

An empirical explanation of the above two loss functions is to find sample points with high confidence that are sufficient to determine all the DoFs of the object pose while keeping the number of sample points small.
It is worth mentioning while there might be multiple potential outputs that will lead to the two loss functions being small, we only require the network to output one of them.
In practice, we found that using a small set of pseudo ground-truth is sufficient for the \pips estimation network to generate meaningful and repeatable positively incentive points.
The reason might be that different shapes share similar geometric patterns. The learned sampling strategy is generalizable across various objects.

\subsection{Training and Inference}
The training of the proposed method contains three stages, as shown in Figure~\ref{fig:training}.
First, an SO(3)-equivariant convolutional implicit network with dense point sampling
(i.e. the teacher model) is optimized to generate the pseudo ground-truth.
Second, the \pips estimation network (i.e. the student model) is trained based on the generated pseudo ground-truth.
Third, another SO(3)-equivariant convolutional implicit network is trained with the sample points generated by the \pips estimation network.

\ys{Despite the relatively high computational cost during the pseudo ground-truth generation procedure, this process is optimized before the neural implicit field with SO(3)-equivariant convolutions is trained.}
Moreover, an interesting phenomenon we observe is that a small number of pseudo ground-truth is sufficient for the \pips estimation network, showing that the trained SO(3)-equivariant convolutional implicit network is effective and has good cross-instance and cross-category generalities.
Once the pseudo ground-truth is generated, the \pips estimation network could be efficiently trained, facilitating the efficient training of downstream networks, e.g. the SO(3)-equivariant convolutional implicit network.

During inference, given an input point cloud, our method generates the positive-incentive sample points by the \pips estimation network and predicts the per-point canonical coordinate on those points. The object pose is then computed from the estimated canonical coordinates by a modified Umeyama algorithm~\cite{umeyama1991least} considering the anisotropic uncertainty.

\subsection{Implementation details}
We implement our method with PyTorch. The networks are trained with the Adam optimizer on a single A100 GPU. The learning rate is $10^{-3}$. For category-level pose estimation, we train separate networks for each category. For the instance-level pose estimation, we train separate networks for each instance.
The training of the generating pseudo ground-truth, \pipsc estimation, and \pipsc estimation components take about $6$, $2$, and $5$ hours, respectively.
The input point cloud includes $1,024$ points. In the \pipsc estimation component, we set the voxel size $h$ as $8$.
In the \pipss estimation component, we set the target sparsity $\rho$ as $0.1$.
When generating the pseudo ground-truth, we set the threshold $\omega$ to $0.5$.
The dense sampling points used to train the teacher model are randomly sampled within a cube in the camera coordinate system. The cube is centered at the center of the input point cloud $P$. The cube's side length is twice the diagonal length of the bounding box of the instance or the categorical mean shape. A total of $4,096$ points are sampled in the cube. In the SO(3)-equivariant convolutional implicit network, the farthest sampling algorithm was employed to downsample the input point cloud. Each kernel contains $13$ support points; the receptive field includes $10$ points.
We add a small random noise to each of the \pipsc and \pipss sample points when feeding them into the SO(3)-equivariant convolutional implicit network.

\section{Results and Evaluation}
\label{sec:results}
We conduct comprehensive experiments to evaluate the proposed method.
First, we provide the experimental results on category-level pose estimation (Sec.~\ref{sec:cat_pose}) and instance-level pose estimation (Sec.~\ref{sec:ins_pose}).
Second, ablation and parameter studies (Sec.~\ref{sec:ablation}) are conducted to analyze the crucial components and parameters of the proposed method.
Third, we conduct a pressure test to quantitatively evaluate the robustness of the proposed method (Sec.~\ref{sec:robustness}).
Last, we show the cross-task generality by demonstrating that the learned sampling strategy applies to other relevant tasks (Sec.~\ref{sec:generality}).

\subsection{Evaluation on Category-level Pose Estimation}
\label{sec:cat_pose}
\subsubsection{Experimental setting}
Category-level pose estimation needs to consider novel objects that have not been trained, so the sample points should be carefully generated to cover all the underlying shapes.
We compare our method with existing methods on category-level pose estimation to show the effects.

The experiments are conducted on two datasets: NOCS-REAL275 and ShapeNet-C.
The NOCS-REAL275 dataset~\cite{wang2019normalized} is a dataset containing $4.3$k training RGB-D images and $2.75$k testing RGB-D images captured from multiple real-world scenes.
The objects in the dataset belong to $6$ categories: bottle, bowl, can, camera, laptop, and mug.
Note that, to better evaluate the learning ability on small sets of training data, we did not use the CAMERA training set in~\cite{wang2019normalized}, which contains a large amount of synthetic data, during training.

\begin{figure}[!t] \centering
\begin{overpic}[width=1.0\linewidth,tics=10]{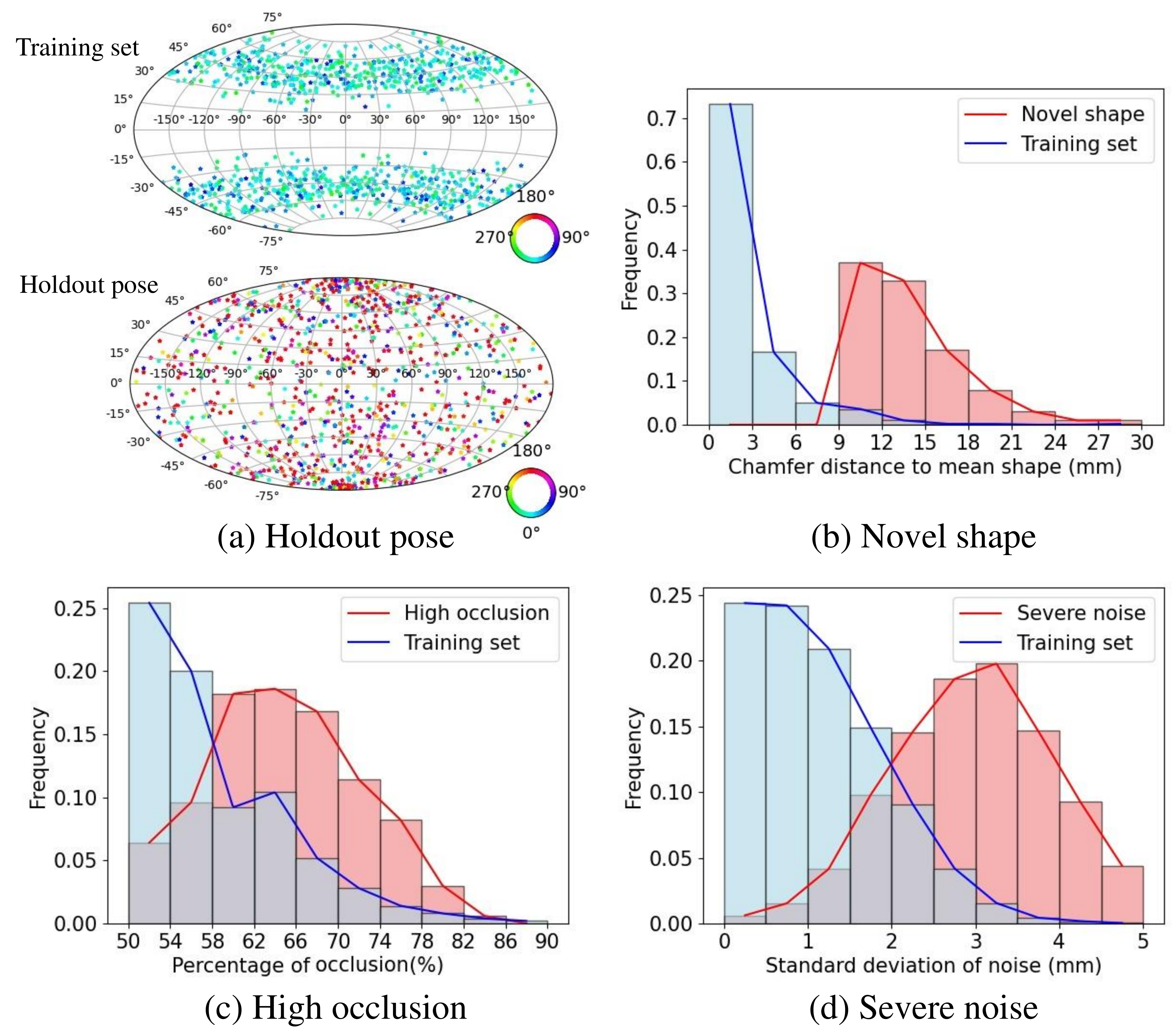}%,grid
\end{overpic}
\caption{The statistical comparisons between the training set and the test set of ShapeNet-C.
}
\vspace{5pt}
\label{fig:dataset}
\end{figure}

To better evaluate the performance of our method in challenging scenarios. We propose a new dataset, i.e., the ShapeNet-C dataset. ShapeNet-C contains $60$k training depth images and $6$k testing depth images. Each RGB-D image contains one object, so no object detection method is needed, and the performance on pose estimation could be better reported.
We set the diagonal length of the bounding box of each object as $1m$.
The dataset contains object categories with large shape variations, such as airplane, chair, and sofa.
\ys{More detailed statistical comparisons of ShapeNet-C to existing datasets are provided in Table~\ref{tab:shapenet-c}. It shows that ShapeNet-C is more challenging in terms of shape diversity and occlusion.}
In particular, the test set of the ShapeNet-C dataset contains the following challenging data:
\begin{compactitem}
\item \textbf{Holdout pose}: objects scanned from untrained camera views.
    We show the distribution of camera locations w.r.t. the object in Figure~\ref{fig:dataset}a, where the color of each point indicates the rotation of the camera along the roll axis.
    \ys{We see a large difference in the distribution of the training and testing samples.}
\item \textbf{Novel shape}: objects with large shape variation from those in the training data. We show the distribution of chamfer distance between each object to the categorical meaning shape~\cite{tian2020shape} in Figure~\ref{fig:dataset}b.
    \ys{We see that 98\% of the samples have a chamfer distance $>$9mm to the category’s mean shape.}
\item \textbf{High occlusion}: highly occluded objects with only a small proportion being observed. The distribution of occlusion percentage is given in Figure~\ref{fig:dataset}c.
    \ys{The statistics show that 91\% of the samples have an occlusion rate exceeding 50\%.}
\item \textbf{Severe noise}: objects with severe scan noise. We add random Gaussian noise to the point cloud. The distribution of the standard deviation is reported in Figure~\ref{fig:dataset}d.
    \ys{It tells that 79\% of the samples have a noise magnitude exceeding 2mm.}
\end{compactitem}
We see that the data distributions of the four subsets are quite different from that of the training set. This makes pose estimation on ShapeNet-C extremely challenging.

We use standard metrics to evaluate the performance on the two datasets, respectively. For NOCS-REAL275, we adopt the intersection over union (IoU) under pre-given thresholds, and the average precision of detected instances for which the error is less than $n^{\circ}$ for rotation and $m$ for translation.
These metrics are utilized to evaluate the performance of object detection and pose estimation, respectively.
For ShapeNet-C, we report the rotational error, and the translational error in the form of mean, and median values. We also report the average precision of instances for which the error is less than $5^{\circ}$ for rotation and $5$cm for translation.

\begin{table}[!t]
	\renewcommand{\arraystretch}{1.1}
	\centering
	\caption{\ys{The comparisons of ShapeNet-C to existing datasets.}}
	\resizebox{\columnwidth}{!}{%
		\begin{tabular}{cccc}
			\hline
			Dataset &  \makecell[c]{Avg. instances\\per category} & \makecell[c]{Avg. Chamfer distance\\to mean shape (mm)} & \makecell[c]{Avg.\\occlusion (\%)}  \\
			\hline
			NOCS-REAL275~\cite{wang2019normalized}   & 4    & 2.7 & 46 \\
			NOCS-CAMERA25~\cite{wang2019normalized}  & 31   & 2.4 & 44 \\
			Wild6D~\cite{Wild6d}        & 360  & 2.2 & 42 \\
			HouseCat6D~\cite{HouseCat6D}     & 19   &  2.8 & 49 \\
			\textbf{ShapeNet-C}    & \textbf{2000} &   \textbf{3.1} & \textbf{53} \\
			\hline
	\end{tabular}}
    \vspace{5pt}
	\label{tab:shapenet-c}
\end{table}

\subsubsection{Quantitative comparison}

\noindent\textbf{Performance on NOCS-REAL275.}\
We first compare our method to the state-of-the-arts on NOCS-REAL275. The baselines include those with different input modalities and those trained on real data or real \& synthetic data.
The quantitative comparisons are given in Table~\ref{tab:nocs_real}.
It shows that our method outperforms all the baselines on all evaluation metrics.
The advantages come from our design of SO(3)-equivariant networks and novel point sampling strategy.

\noindent\textbf{Performance on ShapeNet-C.}\
We then compare our method to several representative baselines on ShapeNet-C.
The quantitative comparisons are reported in Table~\ref{tab:shapenet}. It shows that our method achieves state-of-the-art performance.

%To further analyze the performance, we show the detailed comparisons to a baseline method (i.e., VI-Net~\cite{lin2023vinet}) on the four subsets in Figure~\ref{fig:pressure}. In general, our method outperforms the baseline method in all subsets. There are several phenomena we can observe. First, our method achieves better performance on \emph{holdout pose}, revealing the necessity of the SO(3)-equivariant convolutional implicit network. Second, the better performance of our method on \emph{novel shape} demonstrates the advantage of balancing the exploitation and exploration of the proposed \pips. Last, the results on \emph{high occlusion} and \emph{severe noise} imply the robustness of our method, thanks to the mechanism of sampling in the unseen regions.

\begin{figure*}[!t] \centering
\begin{overpic}[width=0.99\linewidth,tics=10]{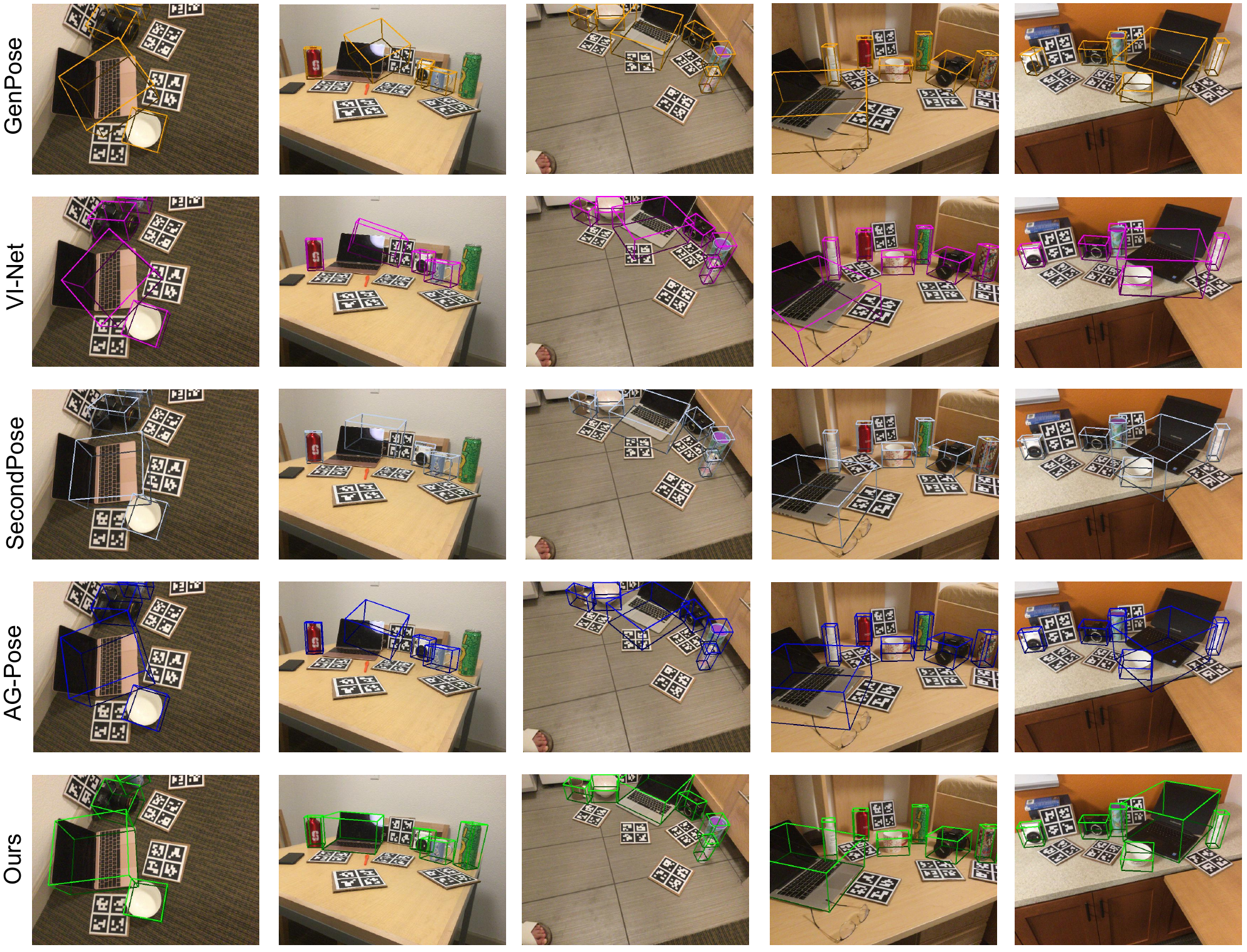}%,grid
\end{overpic}
%\captionsetup{labelfont={color=blue}}
\caption{\ys{The qualitative comparisons on the NOCS-REAL275 dataset. Our method could accurately estimate the pose on all the tested objects, while the baselines cannot.}}
\label{fig:nocs_qual}
\end{figure*}

\begin{table*}[!t]
	\renewcommand{\arraystretch}{1.3}
	\centering
	\caption{\ys{Quantitative comparisons on the NOCS-REAL275 dataset. `*' denotes the
		IoU metrics computed as in~\cite{CATRE}.}}
\resizebox{0.95\linewidth}{!}{
	\begin{tabular}{cccccccccc}
		\hline
		Methods & Input & Data source & Shape Prior & IoU75$^{\ast}$$\uparrow$
		&$5^{\circ}$2cm$\uparrow$ & $5^{\circ}$5cm$\uparrow$ & $10^{\circ}$2cm$\uparrow$ & $10^{\circ}$5cm$\uparrow$ \\
		\hline

		SGPA~\cite{chen2021sgpa} & RGB-D & Syn.+Real & \checkmark &0.37& 0.36 & 0.40 & 0.61 & 0.71   \\
		DPDN~\cite{lin2022category} & RGB-D & Syn.+Real & \checkmark & 0.54& 0.46 & 0.51 & 0.70 & 0.78   \\
		%MH6D~\cite{MH6D} & RGB-D & Syn.+Real  & \checkmark& 0.54 &  0.53 &
		%0.61 & 0.72 & 0.82 \\
		HS-Pose~\cite{HS-Pose} & D & Real  & $\times$ & 0.39 &  0.45 &
		0.55 & 0.69 & 0.84 \\

		GenPose~\cite{yao2023genpose}& D & Real & $\times$ & 0.50 & 0.52 & 0.61 & 0.72 & 0.84  \\
		VI-Net~\cite{lin2023vinet}& RGB-D & Real & $\times$ & 0.48 & 0.50 & 0.58 & 0.71 & 0.82  \\
			SecondPose~\cite{SecondPose}&RGB-D & Syn.+Real & $\times$ & 0.50 & 0.56 & 0.64 & 0.75 & \textbf{0.86}  \\
				AG-Pose~\cite{AGPOSE}& RGB-D & Syn.+Real & $\times$ & 0.61 & 0.57 & 0.65 & 0.75 & 0.85  \\
		\hline
		Ours  & RGB-D & Real  & $\times$ & \textbf{0.63} & \textbf{0.63}  & \textbf{0.68} & \textbf{0.78} & \textbf{0.86}  \\
		\hline
	\end{tabular}
}
	\label{tab:nocs_real}
\end{table*}

\begin{table*}[!t]
	\renewcommand{\arraystretch}{1.3}
	\centering
	\caption{\ys{Quantitative comparisons on the ShapeNet-C dataset.}}
\resizebox{0.95\linewidth}{!}{
	\begin{tabular}{cccccc|ccc}
		\hline
		\multirow{2}{*}{Methods} & \multirow{2}{*}{Input} &\multirow{2}{*}{Data source} & \multicolumn{3}{c|}{Rotation} & \multicolumn{3}{c}{Translation} \\
		\cline{4-9}
		& & &Mean($^{\circ}$) $\downarrow$ & Median($^{\circ}$)$\downarrow$ & $5^{\circ}$ $\uparrow$ & Mean(cm)$\downarrow$ & Median(cm)$\downarrow$ & $5^{\circ}5$cm $\uparrow$ \\
		\hline
		NOCS~\cite{wang2019normalized} & RGB & Syn. & 53.15 & 27.68 & 0.34 & 6.87 & 5.03 & 0.29 \\

		GPV-Pose~\cite{di2022gpv}&  D & Syn. & 50.95 & 20.62 & 0.38 & 6.31 & 4.82 & 0.33 \\
		GenPose~\cite{yao2023genpose} & D & Syn. & 48.29 & 11.81 & 0.41 & 5.76 & 4.29 & 0.37 \\
		VI-Net~\cite{lin2023vinet} & RGB-D  & Syn. & 46.45 & 9.81 & 0.47 & 5.55 & 3.89 & 0.42 \\
		SecondPose~\cite{SecondPose} & RGB-D  & Syn. & 45.77 & 7.87 & 0.49 & 5.42 & 3.91 & 0.45\\
		AG-Pose~\cite{AGPOSE}  & RGB-D  & Syn. & 40.45 & 7.49 & 0.53 & 5.01 & 3.59 & 0.46 \\
		
		\hline
		
		Ours& RGB-D  & Syn. & \textbf{30.70} & \textbf{3.16} & \textbf{0.66} &\textbf{3.84}& \textbf{2.13} & \textbf{0.62} \\
		\hline
	\end{tabular}
	\label{tab:shapenet}
}
\end{table*}

\begin{figure}[!t] \centering
\begin{overpic}[width=1.0\linewidth,tics=10]{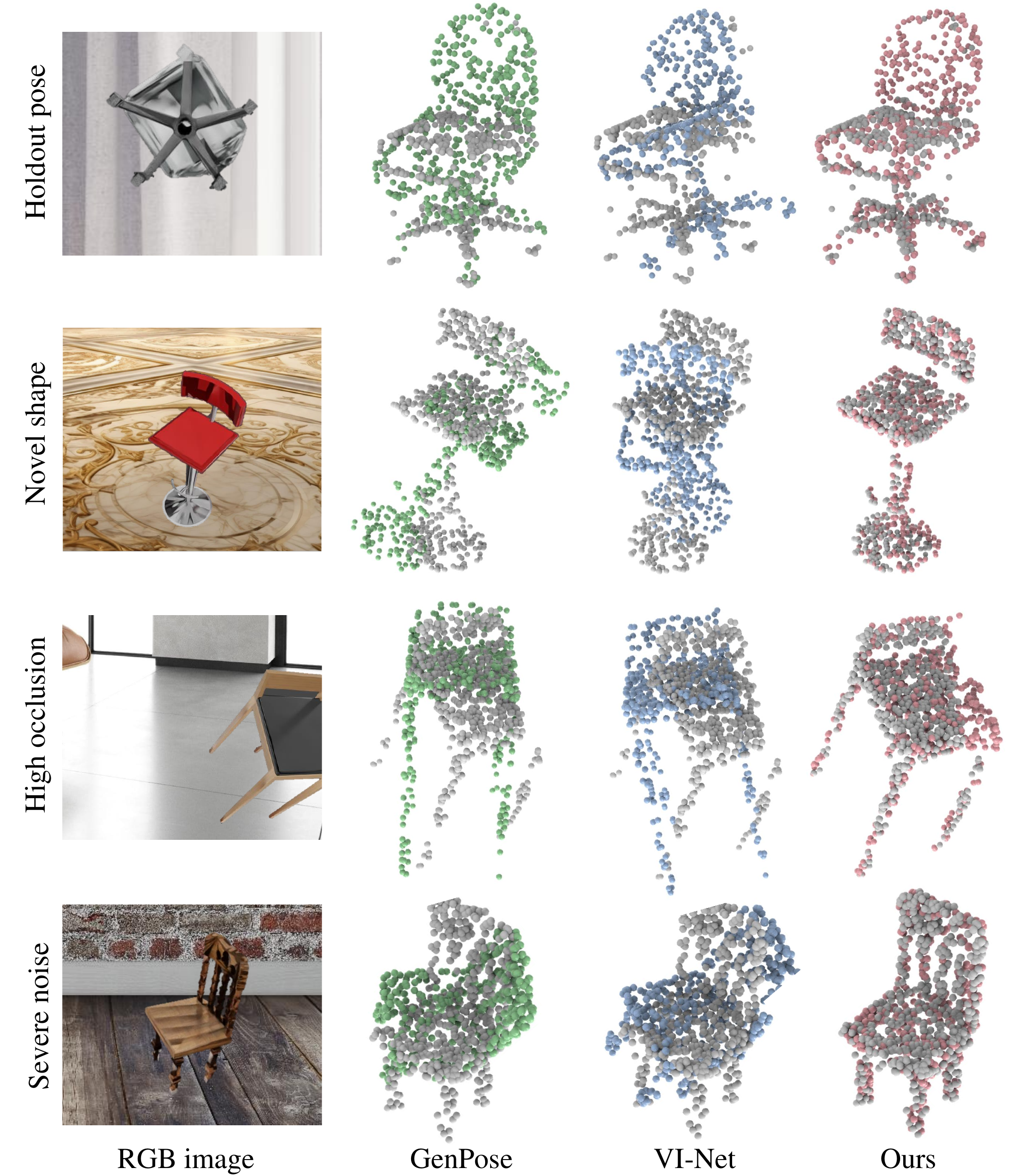}%,grid
\end{overpic}
%\captionsetup{labelfont={color=blue}}
\caption{The qualitative comparisons on the ShapeNet-C dataset. The grey points denote the input points. The colored points represent the canonical shapes transformed with the estimated poses.}
\label{fig:shapenet_qual}
\end{figure}

\subsubsection{Qualitative comparison}
\begin{figure*}[t!] \centering
	\begin{overpic}[width=1.0\linewidth,tics=5]{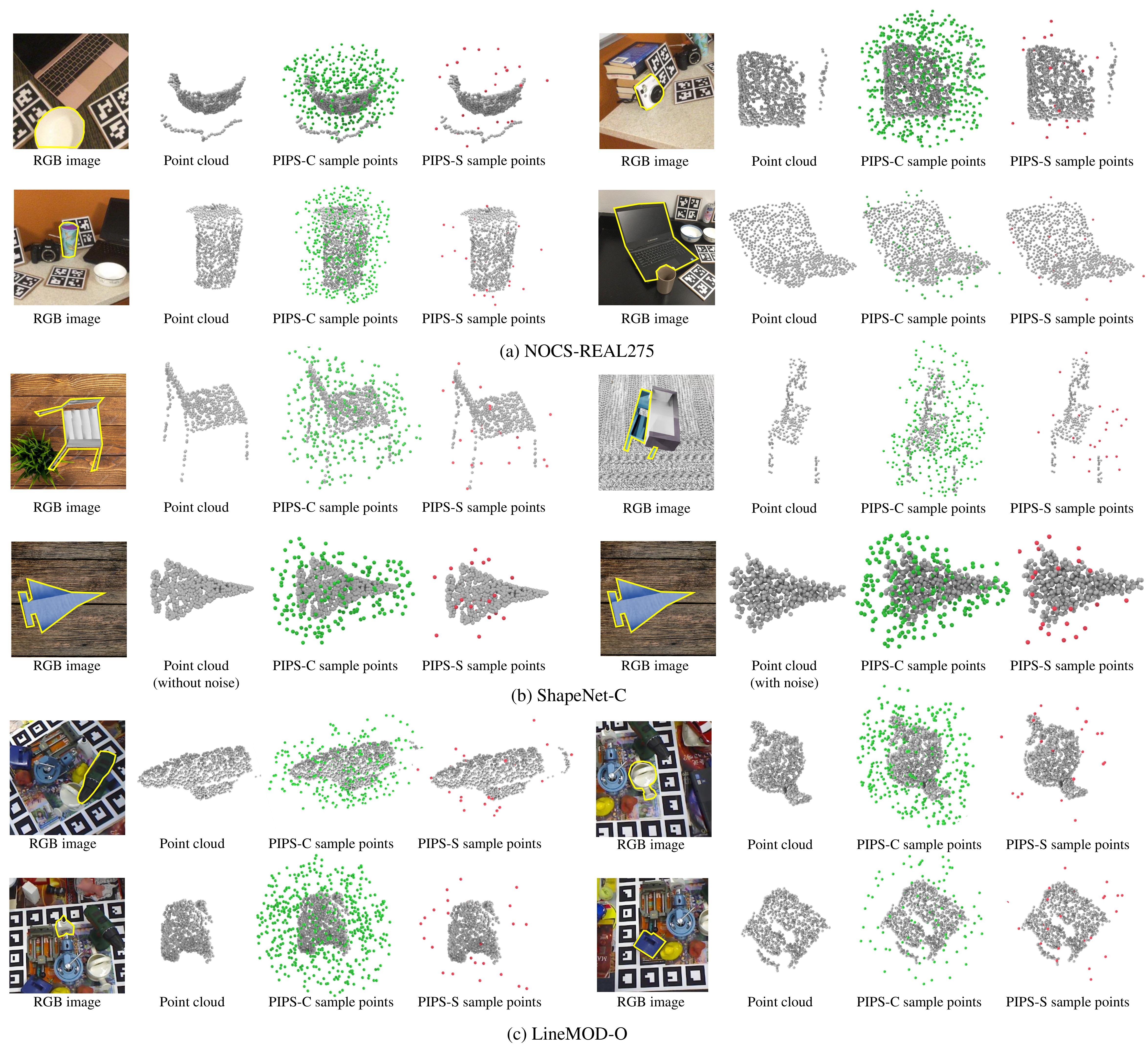}%,grid
    \end{overpic}
   \caption{Visualization of the \pips sample points. The results show that: 1) \pipsc sample points are dense, distributed around the object surface, possibly around the unseen surface region, such as the occluded chair in (b); 2) instead of uniformly distributing around the object surface, the sampling rule varies on different shapes, such as the laptop in (a); 3) \pipss sample points are sparse, covering the crucial regions of the input shape; 4) adding noise to the input points will not substantially change the distribution of \pipss sample points, such as the two cases in last row of (b), where the sampled points are speared near the wings and tail of the airplane despite the noise. The results are produced with only the point cloud as input, without the RGB image features.}
   \label{fig:pips_gallery}
   \vspace{8pt}
\end{figure*}

The qualitative comparisons of the two datasets are visualized in Figure~\ref{fig:nocs_qual} and Figure~\ref{fig:shapenet_qual}. We see that our method could estimate the pose for most objects accurately, while the baselines cannot.
We also provide the visualization of the \pips sample points in Figure~\ref{fig:pips_gallery}a and Figure~\ref{fig:pips_gallery}b.
The results show some interesting phenomena.
First, \pipsc sample points are dense, distributed around the object surface, possibly around the unseen surface region, such as the occluded chair in Figure~\ref{fig:pips_gallery}b.
Second, instead of uniformly distributing around the object surface, the sampling rule varies on different shapes, see the laptop in Figure~\ref{fig:pips_gallery}a.
Third, \pipss sample points are sparse, covering the crucial regions of the input shape.
Last, adding noise to the input points will not substantially change the distribution of \pipss sample points, such as the two cases in the last row of Figure~\ref{fig:pips_gallery}b, where the sampled points are speared near the wings and tail of the airplane despite the noise.

\subsection{Evaluation on Instance-level Pose Estimation}
\label{sec:ins_pose}
\subsubsection{Experimental setting}
We then evaluate our method on instance-level pose estimation on the public dataset LineMOD-O~\cite{brachmann2014learning}.
LineMOD-O is a widely used dataset for 6D object pose estimation of tabletop objects with heavy occlusion.
We use the average recall (AR) as the evaluation metric~\cite{hodan2020bop}.
It is computed as the arithmetic mean of the recall rates for three pose-error functions: the visible surface discrepancy (VSD), the maximum symmetry-aware surface distance (MSSD), and the maximum symmetry-aware projection distance (MSPD).
The recall for each function is considered correct if the estimated error is less than a predefined threshold.

\subsubsection{Quantitative comparison}
The quantitative comparison of our method to baselines is reported in Table~\ref{tab:lmo}.
The baselines include state-of-the-art methods.
In the table, AR is the average recall, i.e., the arithmetic mean of the recall rates for the pose-error functions.
\ys{We see that our method achieves the best performance among all baselines without a refinement procedure.
It is slightly inferior to GPose in the AR metric and has an obvious advantage in computational efficiency.
The good performance of GPose stems from its mechanism of rendering and comparing for pose refinement, which incurs a significant computational cost.}

\subsubsection{Qualitative comparison}
\begin{figure}[t] \centering
\begin{overpic}[width=1.0\linewidth,tics=10]{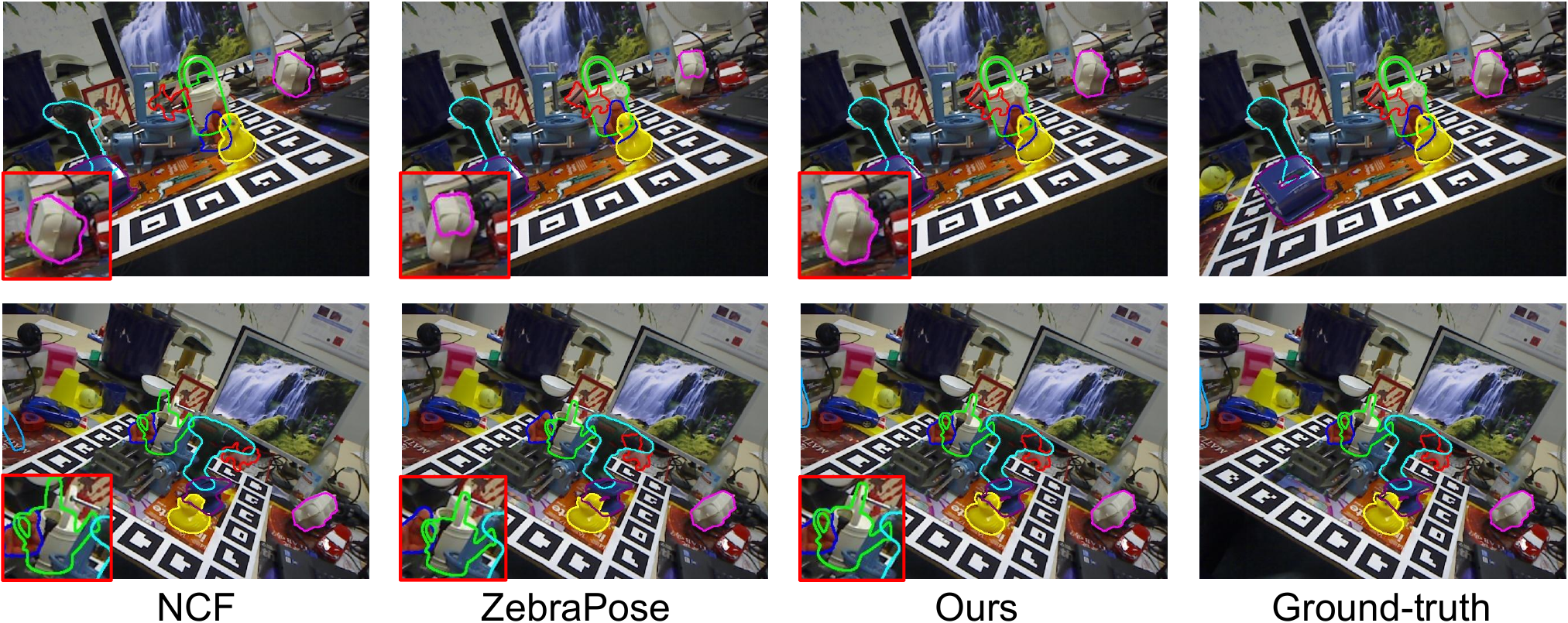}%,grid
\end{overpic}
%\captionsetup{labelfont={color=blue}}
\caption{\ys{The qualitative comparisons on the LineMOD-O dataset. The boundary of the shape transformed by the estimated pose is shown in the figure. Please pay attention to the highlighted objects.}}
\label{fig:lmo_qual}
\vspace{5pt}
\end{figure}

We visualize the comparison of the estimated pose by our method on the LineMOD-O dataset in Figure~\ref{fig:lmo_qual}.
It is clear that our method would lead to more accurate pose estimation in challenging scenarios, as highlighted.
We also provide visualizations of the sample points generated by \pips in Figure~\ref{fig:pips_gallery}c.

\subsection{Ablation and Parameter Studies}
\label{sec:ablation}

In Table~\ref{tab:ablation}, we study the key components of our method to quantify their efficacy.
In Table~\ref{tab:voxel}, Table~\ref{tab:sparsity}, and Table~\ref{tab:threshold}, we study the impact of several parameter settings. The experiments are conducted on the ShapeNet-C dataset.

\noindent\textbf{No \pips and its components.}\
The \pips estimation network, including the \pipsc and \pipss estimation components, is our core contribution. To evaluate the necessity, we turn off the components and retrain the networks. Several conclusions can be drawn from the results. First, the baseline of w/o \pips uses near-surface sampling instead of \pips. It is inferior to our full method, confirming the necessity of detecting positive-incentive sample points.
However, this baseline achieves better performance when compared to state-of-the-art implicit neuron networks in pose estimation~\cite{wan2023socs} ($0.45$ in $5^{\circ}5$cm), demonstrating the effects of the proposed SO(3)-equivariant convolutional implicit network.
Second, the baseline of w/o \pipsc does not filter sample points with low estimation certainty, resulting in a substantial performance drop.
Third, the baseline of w/o \pipss did not consider the geometric stability. Its performance is slightly inferior to our full method.
The results demonstrate the need for the \pips and its two components.

\begin{table}[!t]
	\renewcommand{\arraystretch}{1.3}
	\centering
	\caption{\ys{Quantitative results on the LineMOD-O dataset. The best results are in bold, and the second-best results are underlined.}}
\resizebox{0.9\linewidth}{!}{
\begin{tabular}{ccccc}
	\hline Methods & Data type   & Refinement  & AR & time \\
	\hline

	CDPNv2~\cite{CDPNv2} & RGB  & $\times$ & 62.4 & 0.98 \\
	NCF~\cite{NCF} & RGB  & $\times$ & 63.2 & 7.17 \\
	
	\ys{ZebraPose~\cite{ZebraPose}} & RGB  &  $\times$  &   72.1 & \textbf{0.25} \\
	CosyPose~\cite{cosypose} & RGB-D  & \checkmark & 71.4 & 13.74  \\
	\ys{GDRNPP~\cite{GDRNPP}} & RGB-D  & $\times$ & 71.3 & \underline{0.28}  \\
	\ys{GDRNPP~\cite{GDRNPP} (GPose)} & RGB-D  & \checkmark & \textbf{80.5} & 4.58  \\
	\hline
	
	\ys{Ours} & RGB-D  &  $\times$ & \underline{77.3} &0.39 \\
	\hline
\end{tabular}
	\label{tab:lmo}
    \vspace{10pt}
}
\end{table}

%!TEX root = ../sceneparse.tex

\noindent\textbf{No anisotropic variance.}\
To evaluate the effectiveness of the anisotropic variance in generating the pseudo ground-truth, we replace it with a scalar variance and train the network in Section~\ref{sec:generate_gt} with the loss function by formulation~\ref{eq:adv}. Consequently, the \pipss estimation component is also turned off.
The degraded performance of this baseline validates our design.

\noindent\textbf{No SO(3)-equivariance.}\
The SO(3)-equivariant convolutional implicit network is one of the main contributions.
We replace it with an existing non-equivariant network, i.e., 3D-GCN~\cite{lin2020convolution}.
We see that this baseline is less capable on the ShapeNet-C dataset. We also found that this baseline is significantly inferior to our full method in the holdout pose subset, where our full method achieves $0.61$ in $5^{\circ}5$cm and the baseline of w/o SO(3)-equivariance achieves $0.38$ in $5^{\circ}5$cm.

\ys{
\noindent\textbf{No RGB feature.}\
Adding RGB features seems straightforward, which could enhance the overall performance. To evaluate its importance, we ablate the RGB features in our method and retrain the networks. The performance drop shows that the RGB features are crucial in our method.
}

\noindent\textbf{Parameter settings.}\
Several key parameters are crucial to our method. Here, we evaluate other possibilities to study the rationality of the parameter settings.
The parameters include: 1) the voxel size $h$ of the network output in the \pipsc estimation component; 2) the target sparsity $\rho$ in the training loss function of the \pipss estimation component; 3) the threshold $\omega$ to determine the positive samples in generating pseudo ground-truth for the \pips estimation network.
In general, the results show our method is not very sensitive to these parameter settings.
%Our method achieves the best performance with $h=64$, $\rho=0.1$, and $\omega=0.5$.
Nevertheless, using a large $h$ increases the training efforts and does not lead to significant improvements.
Selecting a small $\rho$ makes the generated sample points sparse, probably missing the detailed information.
Adopting a small $\omega$ incurs uncertain points that are unbeneficial to the network training.

\begin{table}[!t]
	\renewcommand{\arraystretch}{1.1}
	\centering
	\caption{\ys{Ablation study of our method on ShapeNet-C.}}
	\resizebox{\columnwidth}{!}{%
		\begin{tabular}{cccc}
			\hline
			Methods & Median($^{\circ}$)$\downarrow$ & Median(cm)$\downarrow$  &$5^{\circ}$5cm$\uparrow$   \\
		\hline
			w/o  \pips   & 5.92 & 2.78 & 0.51  \\
			w/o  \pipsc   & 5.34 & 2.61 & 0.53  \\
			w/o  \pipss  & 4.23  & 2.49  & 0.57  \\
			w/o  anisotropic variance   & 4.51 & 2.89 & 0.55  \\
			w/o SO(3)-equivariance   & 5.18 & 3.02 & 0.49 \\
            \ys{w/o RGB}   & 3.48 & 2.67 & 0.59 \\							
			\hline
			\ys{Full method}  & \textbf{3.16} & \textbf{2.13} & \textbf{0.62} \\
			\hline

	\end{tabular}}
	\label{tab:ablation}
\end{table}

\begin{figure}[t] \centering
\begin{overpic}[width=0.9\linewidth,tics=10]{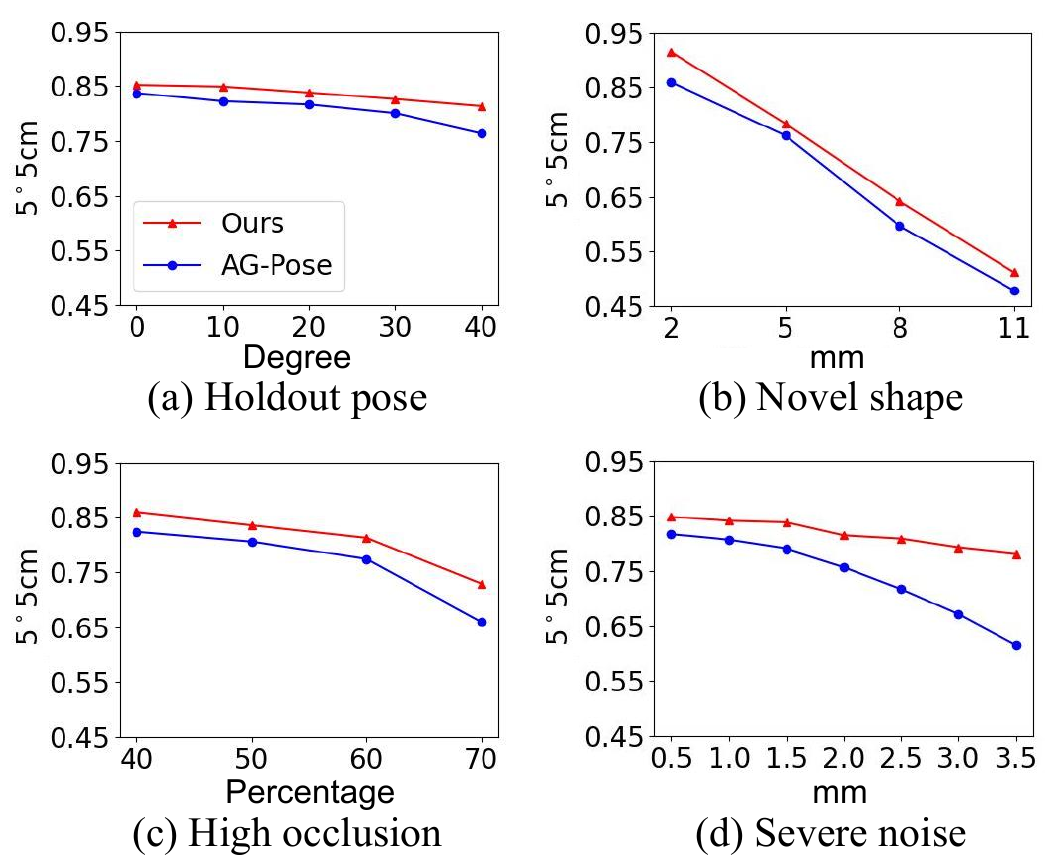}%,grid
\end{overpic}
%\captionsetup{labelfont={color=blue}}
\caption{\ys{Robustness evaluation. The quantitative comparisons of our method to a baseline method on the four challenging subsets of the NOCS-CAMERA25 dataset. Our method is more robust on unseen poses, novel shapes, high occlusion, and severe noise.}}
\label{fig:pressure}
\end{figure}

\subsection{Robustness Evaluation}
\label{sec:robustness}

\noindent\textbf{Robustness in challenging scenarios.}\
\ys{To evaluate the robustness of our method, we conduct a pressure test on the NOCS-CAMERA25 dataset following a similar data division in Section~\ref{sec:cat_pose}. The quantitative comparisons to the baseline method AG-Pose~\cite{AGPOSE} are shown in Figure~\ref{fig:pressure}.
The plots show that our method is more stable and robust in scenarios with holdout pose, novel shape, high occlusion, and severe noise.
There are several phenomena we can observe. First, our method achieves better performance on \emph{holdout pose}, revealing the necessity of the SO(3)-equivariant convolutional implicit network. Second, the better performance of our method on \emph{novel shape} demonstrates the advantage of balancing the exploitation and exploration of the proposed \pips. Last, the results on \emph{high occlusion} and \emph{severe noise} imply the robustness of our method, thanks to the mechanism of sampling in the unobserved regions.}

\noindent\textbf{Robustness to perturbation on sample points.}\
Another interesting problem is the method's robustness under perturbations of the \pipss sample points.
To investigate this problem, we randomly add Gaussian noise to the generated \pipss sample points during both the training and testing. We found the influence is not significant when the standard deviation of noise is less than $1$ cm.
Moreover, we conducted another experiment that randomly dropped out some \pipss sample points during the training of testing of the student model. Results show the performance is robust with about $30\%$ of the sample points being turned off.

\begin{table}[!t]
	\renewcommand{\arraystretch}{1.3}
	\centering
	\caption{\ys{Effect of voxel size $h$ in \pipsc estimation.}}
\resizebox{1.0\linewidth}{!}{
	\begin{tabular}{ccccc}
		\hline
		Voxel size & Median($^{\circ}$)$\downarrow$ & Median(cm)$\downarrow$  & $5^{\circ}$ $\uparrow$ &$5^{\circ}$5cm$\uparrow$ \\
		\hline
		4 & 5.42 & 3.15 & 0.56 & 0.49 \\
		8 & \textbf{3.16} & 2.13 & \textbf{0.68} & \textbf{0.64}  \\
		16 & 3.28 & \textbf{2.08} & \textbf{0.68} & 0.62 \\
		32 & 3.71 & 2.34 & 0.63 & 0.58 \\
		\hline	
	\end{tabular}
}
	\label{tab:voxel}
\end{table}

\begin{table}[!t]
	\renewcommand{\arraystretch}{1.3}
	\centering
	\caption{\ys{Effect of target sparsity $\rho$ in \pipss estimation.}}
\resizebox{1.0\linewidth}{!}{
	\begin{tabular}{ccccc}
		\hline
		Target sparsity & Median($^{\circ}$)$\downarrow$ & Median(cm)$\downarrow$  & $5^{\circ}$ $\uparrow$ &$5^{\circ}$5cm$\uparrow$  \\

		\hline
0.3 & 3.42 & 2.65 & 0.63 & 0.56 \\
0.2 & 3.28 & 2.38 & \textbf{0.68} & 0.61 \\
0.1 & \textbf{3.16} & \textbf{2.13} & 0.66 & \textbf{0.62} \\
0.05 & 3.71 & 2.84 & 0.59 & 0.52 \\
\hline
	\end{tabular}
}
	\label{tab:sparsity}
\end{table}

\begin{table}[!t]
	\renewcommand{\arraystretch}{1.3}
	\centering
	\caption{\ys{Effect of threshold $\omega$ in positive sample selection of the \pips estimation network.}}
\resizebox{1.0\linewidth}{!}{
	\begin{tabular}{ccccc}
		\hline
		Threshold  & Median($^{\circ}$)$\downarrow$ & Median(cm)$\downarrow$  & $5^{\circ}$ $\uparrow$ &$5^{\circ}$5cm$\uparrow$  \\
		\hline
		0.9 & 4.32 & 3.24 & 0.57 & 0.49 \\
0.7 & 3.89 & 2.95 & 0.61 & 0.54 \\
0.5 & \textbf{3.16} & \textbf{2.13} & \textbf{0.66} & \textbf{0.62} \\
0.3 & 3.58 & 2.75 & 0.63 & 0.56 \\
		\hline
	\end{tabular}
}
	\label{tab:threshold}
\end{table}

\subsection{Cross-task Generality of \pips}
\label{sec:generality}
\ys{The \pips estimation network essentially learns to generate sample points whose features are informative and representative, allowing for a direction-aware quantification of prediction confidence for 3D coordinates.
Consequently, the learned sampling strategy might be useful to other relevant tasks that involve per-point 3D coordinate regression and require uncertainty estimation for these outputs.}
To show the cross-task generality of \pips, we apply the \pips estimation network trained on category-level pose estimation (denoted as $\pips_\text{CAT}$) to two relevant tasks: instance-level pose estimation and shape reconstruction. Specifically, $\pips_\text{CAT}$ is trained on the ShapeNet-C dataset.

\noindent\textbf{Generalizing to instance-level pose estimation.}\
We first apply $\pips_\text{CAT}$ to the instance-level pose estimation task.
To verify its generality, we train an alternative SO(3)-equivariant convolutional implicit networks to estimate object pose in the LineMOD-O dataset with the \pips sample points generated by $\pips_\text{CAT}$.
We then compare it with the original network trained on instance-level pose estimation (denoted as $\pips_\text{INS}$) as described in Section~\ref{sec:ins_pose}.
Interestingly, we found $\pips_\text{CAT}$ and $\pips_\text{INS}$ produce similar distributions of sample points, resulting in comparable performances of the alternative method.

\begin{figure}[t!]
   \begin{overpic}[width=1.0\linewidth,tics=10]{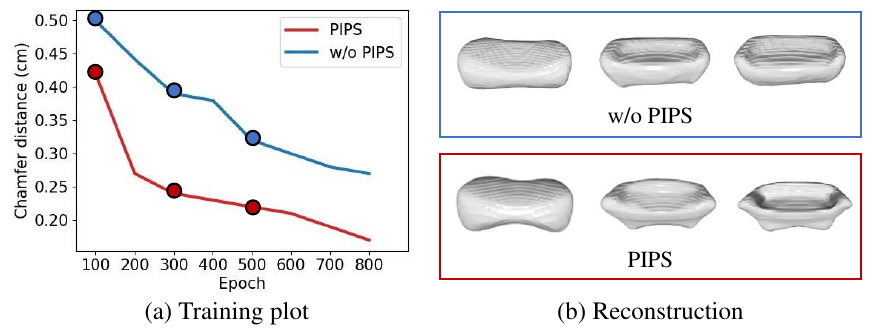}
   \end{overpic}
   \caption{(a) The training plot of the generalizing to shape reconstruction experiment. (b) Examples of the reconstructed shapes during the network training.}
   \label{fig:generality}
\end{figure}

\noindent\textbf{Generalizing to shape reconstruction.}\
To validate the effects of $\pips_\text{CAT}$ in the shape reconstruction task, we train two SO(3)-equivariant convolutional implicit networks to estimate the point-wise SDFs similar to DeepSDF~\cite{park2019deepsdf}.
The networks take single-view depth images as input and learn continuous signed distance functions to represent the complete shape. The object is then reconstructed by the Marching Cubes algorithm~\cite{lorensen1998marching}.
The two networks are trained with random sample points and the \pipsc sample points by $\pips_\text{CAT}$, respectively.
The training plots of the two networks are visualized in Figure~\ref{fig:generality}a, which shows that \pipsc sampled points lead to a faster error drop.
The reconstructed shapes generated by the two methods are visualized in Figure~\ref{fig:generality}b.
The results indicate that the \pipsc trained on pose estimation is indeed helpful to shape reconstruction, demonstrating the cross-task generality.

%!TEX root = ../sceneparse.tex

\section{Conclusion and Discussion}
\label{sec:conclusion}
In this work, we studied the problem of positive-incentive point sampling for neural implicit fields.
To this end, we proposed a \pips estimation network that generates sparse sample points that would gain sufficient information to determine all the DoFs of the object pose by training on them.
Moreover, an SO(3)-equivariant convolutional implicit network that estimates point-level attributes with SO(3)-equivariance at any sample point is developed, outperforming most existing implicit neuron networks in pose estimation.
Our method achieves state-of-the-art performance on three datasets.
Several crucial conclusions can be drawn from this study. First, dense sampling is unnecessary for pose estimation with neural implicit fields. Second, positive-incentive point sampling can be estimated with a learning-based approach. Third, the learned sampling strategy is generalizable to other relevant tasks.

\ys{Our method estimates 3D anisotropic uncertainty from a pointset, making it applicable to tasks that involve per-point 3D coordinate regression and require uncertainty estimation for these outputs. Promising applications include: 1) Localization and Mapping: Our method can be used to select landmarks or points in a point cloud that provide the most reliable information for pose estimation, reducing drift and improving robustness in perceptually challenging environments; 2) Point Cloud Registration: Our method can prioritize point correspondences with high certainty and geometric stability, leading to faster convergence and higher accuracy in alignment algorithms.}

Our method has the following limitations.
First, the \pips estimation network is trained with the pseudo ground-truth generated by a teacher model which requires additional training efforts. Integrating the two networks into a unified framework might make the method concise and further reduce the training costs.
Second, our method cannot handle the problem of pose ambiguity caused by occlusions.
An interesting direction is to opt for a diffusion model-based network that could generate multiple outputs given an input point cloud.
For future work, we expect to apply the proposed \pips to Neural Radiance Fields~\cite{mildenhall2021nerf} and 3D Gaussian Splatting~\cite{kerbl20233d}, which might enable more efficient network training and improve the quality of synthesis images in large-scale scenes. 

\appendices
\setcounter{equation}{0}

\ifCLASSOPTIONcompsoc
  % The Computer Society usually uses the plural form
  \section*{Acknowledgments}
We thank the anonymous reviewers for their valuable comments.
This work was supported by the National Natural Science Foundation of China (62325211, 62132021, T2521006, 62302517), the Natural Science Foundation of Hunan Province (2023JJ20051), the Science and Technology Innovation Program of Hunan Province (2023RC3011), and the Cornerstone Foundation of NUDT (JS24-03).

\else
  % regular IEEE prefers the singular form
  \section*{Acknowledgment}
\fi

\ifCLASSOPTIONcaptionsoff
  \newpage
\fi

\bibliographystyle{ieee_fullname}
\bibliography{egbib}

\begin{IEEEbiography}[{\includegraphics[width=1in,height=1.25in,clip,keepaspectratio]{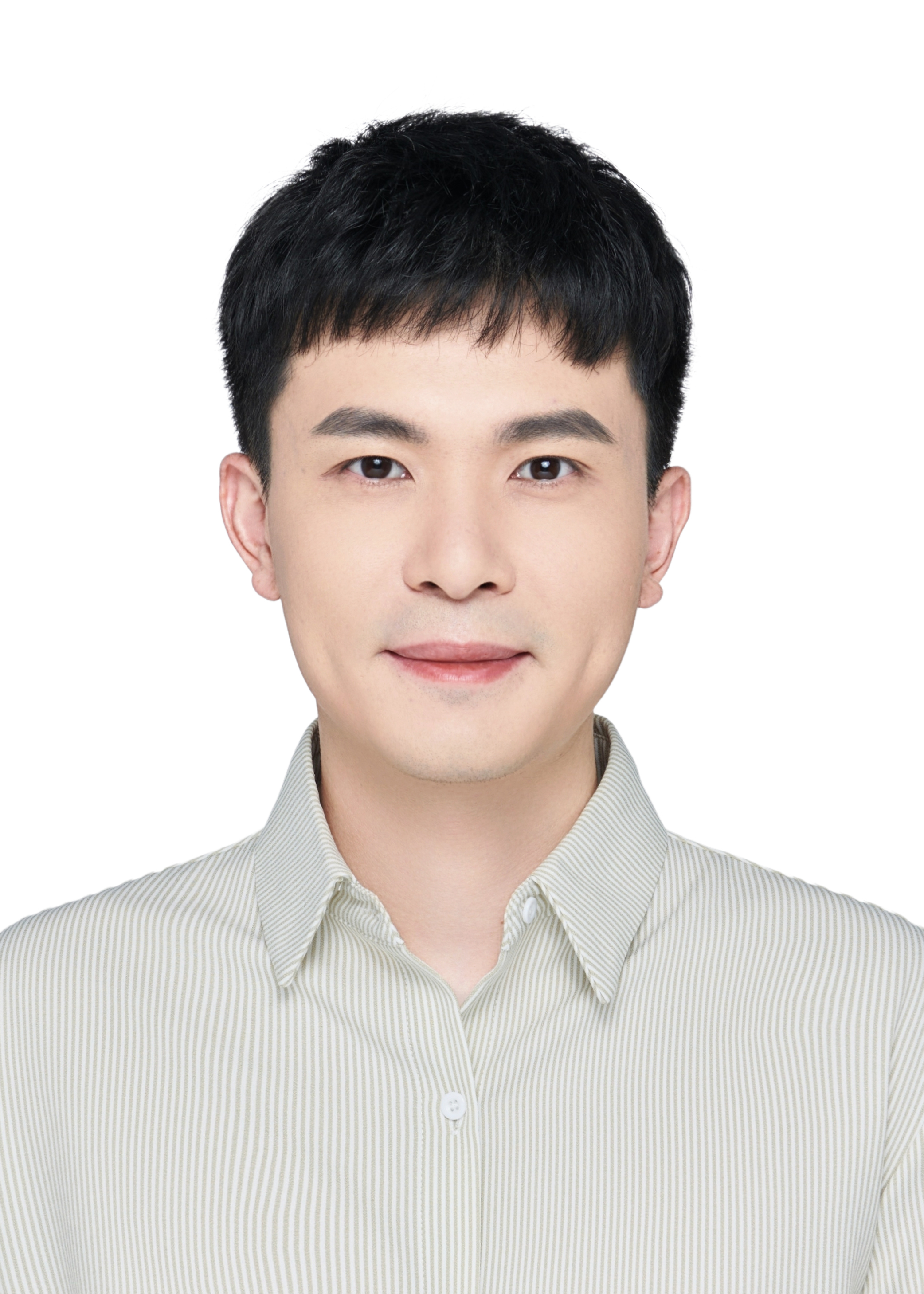}}]{Yifei Shi}
is a Professor at the College of Intelligence Science and Technology, National University of Defense Technology (NUDT). He received his Ph.D. degree in computer science from NUDT in 2019. During 2017-2018, he was a visiting student research collaborator at Princeton University, advised by Thomas Funkhouser and Szymon Rusinkiewicz. His research interests mainly include computer vision, computer graphics, especially on object/scene analysis and manipulation by machine learning and geometric processing techniques. He has published 20+ papers in top-tier conferences and journals, including CVPR, ECCV, ICCV, SIGGRAPH Asia, IEEE Transactions on Pattern Analysis and Machine Intelligence, and ACM Transactions on Graphics.
\end{IEEEbiography}

\begin{IEEEbiography}[{\includegraphics[width=1in,height=1.25in,clip,keepaspectratio]{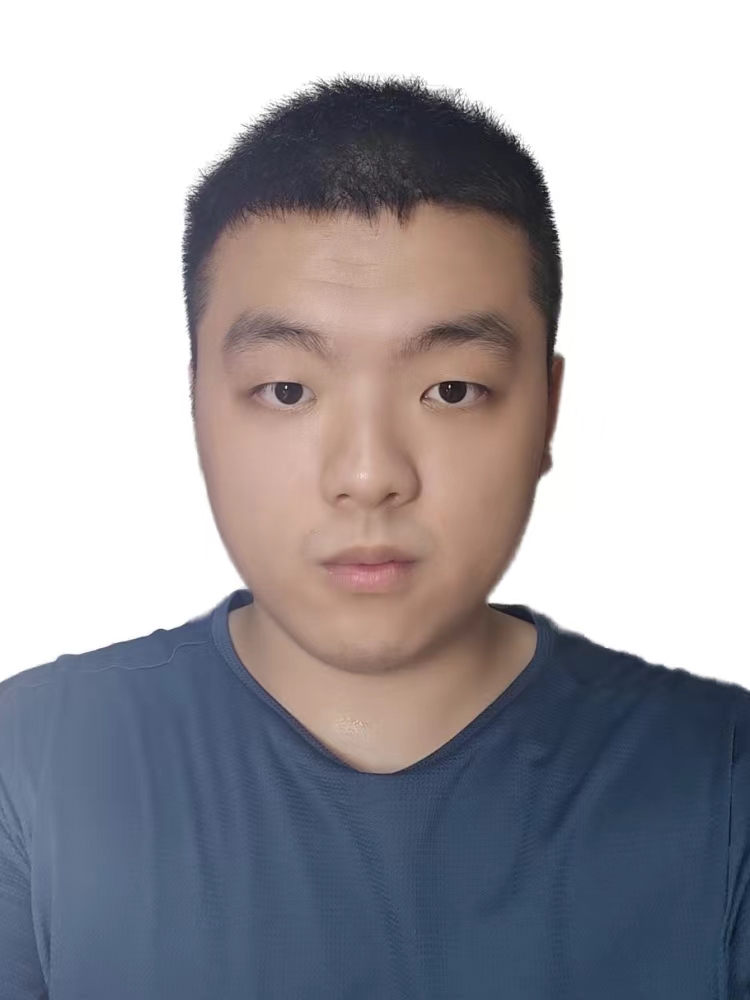}}]{Boyan Wan}
is pursuing his Ph.D. degree in College of Computer Science, NUDT. His research interests include pose estimation, 3D reconstruction and generative model. He has published several papers in international journals and conferences such as the International Conference on Computer Vision, lEEE TASLP. etc.
\end{IEEEbiography}

\begin{IEEEbiography}[{\includegraphics[width=1in,height=1.25in,clip,keepaspectratio]{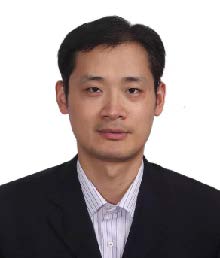}}]{Xin Xu}
received the B.S. degree in electrical engineering from the Department of Automatic Control, National University of Defense Technology (NUDT), Changsha, China, in 1996, and the Ph.D. degree in control science and engineering from the College of Mechatronics and Automation (CMA), NUDT. He has been a Visiting Scientist for cooperation research at The Hong Kong Polytechnic University, the University of Alberta, and the University of Strathclyde. He is currently a Full Professor with the College of Intelligence Science and Technology and the Director of the Department of Artificial Intelligence, NUDT. He has published two monographs and more than 200 articles, with an H-index 34. His main research interests include machine learning and autonomous control of robots and intelligent unmanned systems. He received the Distinguished Young Scholars’ Funds of the National Natural Science Foundation of China in 2018. He was one of the recipients of The George N. Saridis Best Transactions Paper Award of IEEE TRANSACTIONS ON INTELLIGENT TRANSPORTATION SYSTEM, the second-class National Natural Science Award of China, and two first-class Natural Science Awards of Hunan Province, China. He is an Associate Editor of IEEE TRANSACTIONS ON SYSTEM, MAN AND CYBERNETICS: SYSTEMS, Information Sciences, International Journal of Robotics and Automation, the Associate Editor-in-Chief of CAAI Transactions on Intelligence Technology, and an Editorial Board Member of Journal of Control Theory and Applications.
\end{IEEEbiography}

\begin{IEEEbiography}[{\includegraphics[width=1in,height=1.25in,clip,keepaspectratio]{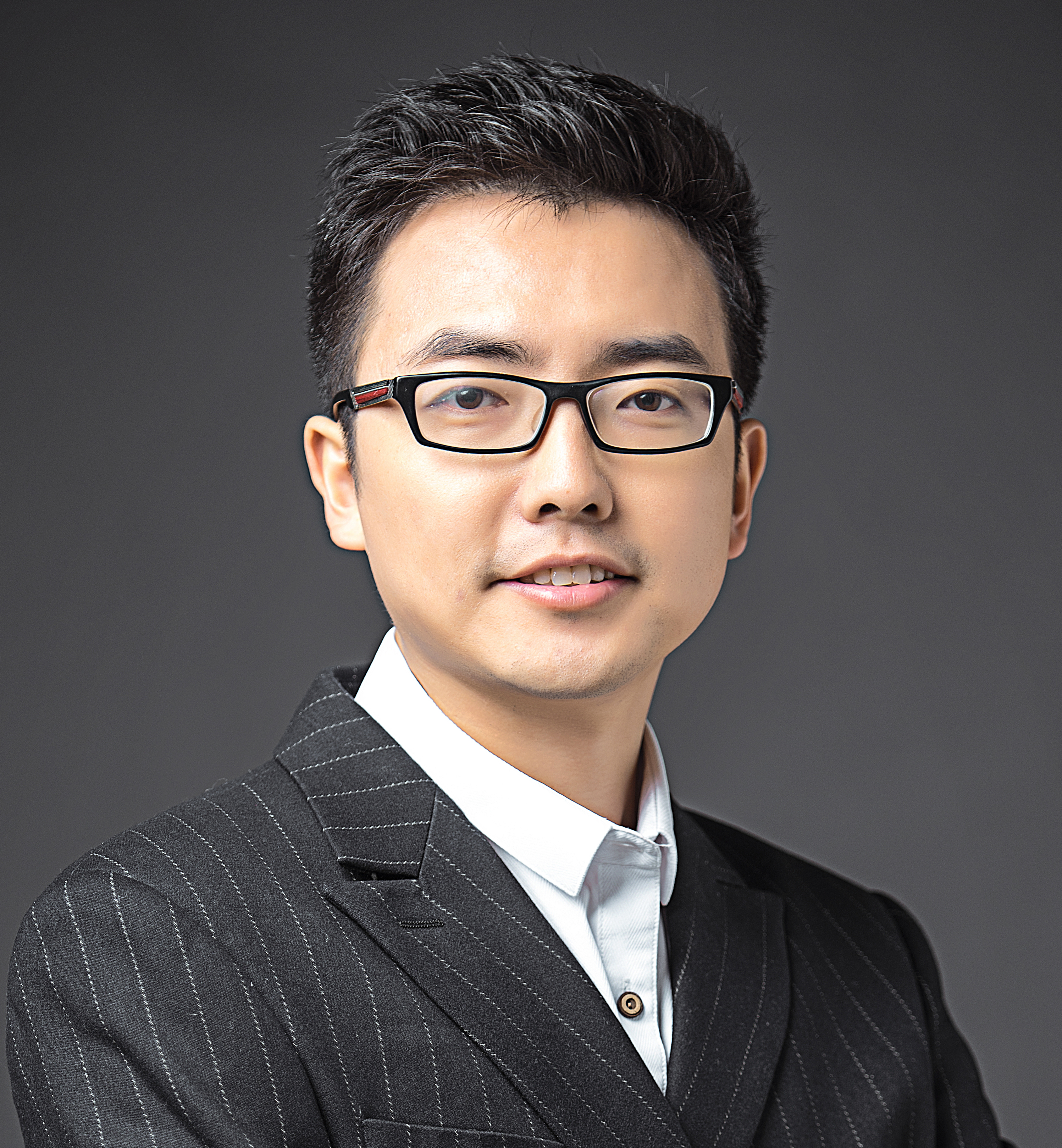}}]{Kai Xu}
is a Professor at the College of Computer, NUDT, where he received his Ph.D. in 2011. He conducted visiting research at Simon Fraser University and Princeton University. His research interests include geometric modeling and shape analysis, especially on data-driven approaches to the problems in those directions, as well as 3D vision and its robotic applications. He has published over 80 research papers, including 20+ SIGGRAPH/TOG papers. He has co-organized several SIGGRAPH Asia courses and Eurographics STAR tutorials. He serves on the editorial board of ACM Transactions on Graphics, Computer Graphics Forum, Computers \& Graphics, and The Visual Computer. He also served as program co-chair of CAD/Graphics 2017, ICVRV 2017 and ISVC 2018, as well as PC member for several prestigious conferences including SIGGRAPH, SIGGRAPH Asia, Eurographics, SGP, PG, etc. His research work can be found in his personal website: www.kevinkaixu.net.
\end{IEEEbiography}

\end{document}